\documentclass[lettersize,journal]{IEEEtran}
\usepackage{amsmath,amsfonts}
\usepackage{algorithm, algpseudocode}
\usepackage{array}
\usepackage[caption=false,font=normalsize,labelfont=sf,textfont=sf]{subfig}
\usepackage{textcomp}
\usepackage{stfloats}
\usepackage{url}
\usepackage{verbatim}
\usepackage{graphicx}
\usepackage{cite}
\usepackage{amsmath,amssymb,amsfonts,bm}
\usepackage{xcolor}
\usepackage{dsfont}
\usepackage{multirow}
\usepackage{booktabs}
\newtheorem{rmk}{Remark}

\newtheorem{Assum}{Assumption}
\newtheorem{Theorem}{Theorem}
\newtheorem{Corollary}{Corollary}

\def\changeBibColor#1{%
	\in@{#1}{}
	\ifin@\color{blue}\else\normalcolor\fi
}

\begin{document}

\title{Exploiting Radio Frequency Fingerprints for Device Identification: Tackling Cross-receiver Challenges in the Source-data-free Scenario}

\author{\IEEEauthorblockN{Liu Yang, Qiang Li, Luxiong Wen and Jian Yang}
	\thanks{L. Yang, Q. Li  and L. Wen   are with
		School of Information and Communication  Engineering, University of Electronic Science and Technology of China, Chengdu, P.~R.~China, 611731.  
		
		J. Yang  is with Beijing Institute
		of Technology, Beijing, P. R. China, 100081, and also with Laboratory of
		Electromagnetic Space Cognition and Intelligent Control, Beijing, P. R. China,
		100083. 
		
		This work was supported by the Natural Science Foundation of China (NSFC) under Grant 62171110.
		
		Q. Li is the corresponding author. E-mail: lq@uestc.edu.cn.}}

\maketitle

\begin{abstract}

With the rapid proliferation of edge computing, Radio Frequency Fingerprint Identification (RFFI) has become increasingly important for secure device authentication. However, practical deployment of deep learning-based RFFI models is hindered by a critical challenge: their performance often degrades significantly when applied across receivers with different hardware characteristics due to distribution shifts introduced by receiver variation. To address this, we investigate the source-data-free cross-receiver RFFI (SCRFFI) problem, where a model pretrained on labeled signals from a source receiver must adapt to unlabeled signals from a target receiver, without access to any source-domain data during adaptation.
We first formulate a novel constrained pseudo-labeling–based SCRFFI adaptation framework, and provide a theoretical analysis of its generalization performance. Our analysis highlights a key insight: the target-domain performance is highly sensitive to the quality of the pseudo-labels generated during adaptation. Motivated by this, we propose Momentum Soft pseudo-label Source Hypothesis Transfer (MS-SHOT), a new method for SCRFFI that incorporates momentum-center-guided soft pseudo-labeling and enforces global structural constraints to encourage confident and diverse predictions. Notably, MS-SHOT effectively addresses scenarios involving label shift or unknown, non-uniform class distributions in the target domain—a significant limitation of prior methods.
Extensive experiments on real-world datasets demonstrate that MS-SHOT consistently outperforms existing approaches in both accuracy and robustness, offering a practical and scalable solution for source-data-free cross-receiver adaptation in RFFI.

\end{abstract}

\begin{IEEEkeywords}
Radio Frequency Fingerprint Identification,  cross-receiver, source-data-free
\end{IEEEkeywords}

\section{Introduction}

With the growing number of connected devices, such as smartphones, Internet of Things (IoT) devices, and embedded systems, ensuring secure access has become an increasingly significant challenge. As the deployment of such devices continues to expand, the demand for efficient and reliable authentication methods has grown accordingly. In edge intelligence systems, more processing and decision-making are being shifted to edge devices, enabling faster responses and greater autonomy. However, this shift also introduces new security challenges. Traditional authentication methods, such as Internet Protocol (IP) or Media Access Control (MAC) addresses, are often vulnerable to spoofing and fail to provide the necessary level of security in these dynamic and resource-constrained environments~\cite{zou2016survey}.

Radio Frequency (RF) fingerprint identification (RFFI) offers a robust alternative by leveraging the unique hardware characteristics of wireless devices for authentication~\cite{wang2016wireless}. These RF fingerprints are intrinsic to the physical hardware, making them highly tamper-resistant and providing a more secure and reliable solution~\cite{shen2021radio, chen2022radio, zhang2023radio}.

\begin{figure}[]
\centerline{\includegraphics[scale=0.4]{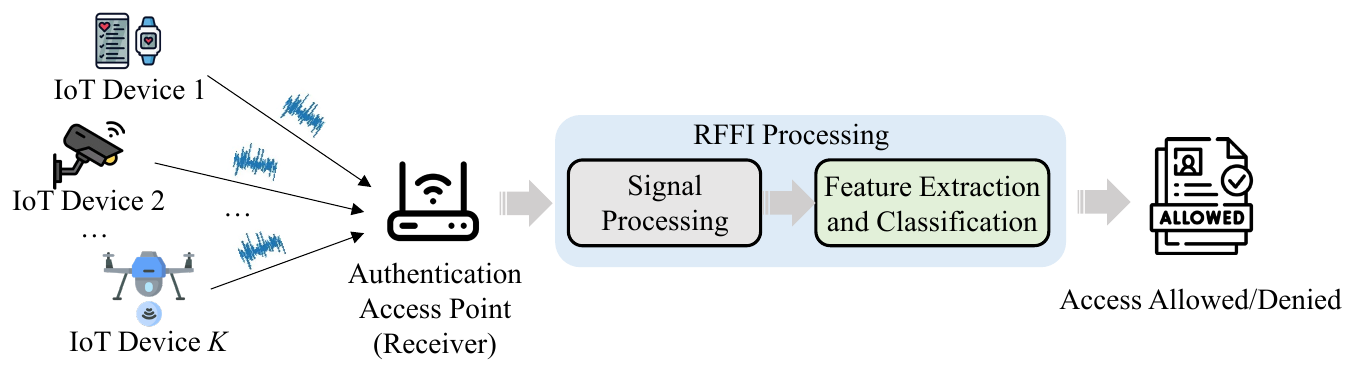}}
\caption{An RFFI-based access authentication system workflow. The diagram shows the workflow of an RFFI-based access authentication system. IoT devices (e.g., smartphones, cameras, drones) transmit wireless signals, which are captured by an authentication access point (receiver). The system performs
	signal processing to filter out noise and enhance signal quality, followed by feature extraction and classification to identify unique RF fingerprints. Based on the classification results, the system decides whether to grant or deny access to the device.}
\label{fig:system}
\vspace{-0.75em}
\end{figure}

Fig.~\ref{fig:system} illustrates a typical RFFI-based access authentication system. In this system, various IoT devices, such as smartphones, surveillance cameras, and drones, transmit wireless signals that are captured by the receiver on the authentication access point. These signals undergo RFFI processing, which consists of two main stages: signal processing and feature extraction and classification. The signal processing stage filters out noise and amplifies the received signal, preparing it for the subsequent analysis. The feature extraction and classification stage identifies the unique RF fingerprints embedded in the signal, which are used to classify the transmitting device and determine whether access should be allowed or denied.
Several studies have been conducted on RFFI, including traditional signal processing and deep learning (DL) approaches. Traditional methods depend on the handcrafted specialized features, such as phase noise, time-frequency characteristics, and channel state information (CSI), to identify devices~\cite{knox2010agc, brik2008wireless, kong2023physical}. Despite showing promising results, these techniques are limited by the requirement for feature engineering expertise, which can restrict their ability to handle complex wireless signals. Moreover, the handcrafted features are difficult to adapt to the growing number of wireless devices~\cite{xie2023disentangled}. 
Recently, deep learning techniques have emerged as promising tools for RFFI. In contrast to traditional methods, DL methods can capture the fine features in the RF signals, resulting in more accurate and robust identification~\cite{merchant2018deep, jagannath2022comprehensive,sankhe2019oracle,yin2021lte, zhang2023radio, shen2023towards, yang2022open}. Specifically, Merchant \emph{et al.}~\cite{merchant2018deep} first adopted Convolutional Neural Network (CNN) for RFFI to extract features automatically from received signals. Sankhe \emph{et al.}~\cite{sankhe2019oracle} proposed a hybrid approach, containing a more versatile distance-based classifier and an automatic feature extractor. Shen \emph{et al.}~\cite{shen2023towards} proposed four neural networks that can process signals of variable lengths and adopt data augmentation during training which can improve the model’s robustness to noise.
DL models do not require manual feature engineering and can learn complex representations of the RF signals automatically, making them more suitable for handling complex wireless signals.

Despite these advances, most DL-based RFFI approaches focus primarily on signal processing and feature extraction, overlooking the critical role of the receiver in shaping signal features. In edge intelligence scenarios, where decentralized systems operate with multiple edge nodes, the receivers at different nodes often have varying hardware characteristics. This variability can significantly impact the performance of RFFI models, as these models are typically trained on signals collected from a specific receiver. When deployed on systems with different receivers, such as those at other edge nodes, the models may struggle to generalize. This is because the received signal's features are influenced not only by the transmitter's hardware but also by the receiver's hardware. Differences between training and deployment receivers lead to a shift in feature distributions, known as domain shift, which can degrade the model's recognition accuracy~\cite{merchant2019toward,shen2022towards,zha2023cross, liu2023receiver, zhao2023gan, yang2023led}; see also the results in Section~\ref{sec:experiment}.

A straightforward solution to mitigate the impact of receiver variation is to retrain the model using data collected from the deployment receiver. However, this approach may be impractical due to the high cost and time involved in labeling the new signals from the deployment receiver. Therefore, the \textit{cross-receiver RFFI problem} can be framed as a scenario where the model must adapt to a new receiver using only labeled signals from the original receiver and unlabeled signals from the new receiver~\cite{zha2023cross}.
To address the cross-receiver RFFI problem, transfer learning techniques \cite{ganin2016domain} are widely applied.  Merchant \emph{et al.}~\cite{merchant2019toward} advocated reducing the effect of receivers by minimizing the distance of the predicted probability from both receivers. Shen \emph{et al.}~\cite{shen2022towards} focused on learning receiver-independent features by adversarial training over multiple receivers’ signals. Zha \emph{et al.}~\cite{zha2023cross} employed unsupervised pre-training through contrastive learning to extract receiver-agnostic features and subsequently optimize the model via subdomain adaptation to enhance identification performance. Liu \emph{et al.}~\cite{liu2023receiver} proposed a receiver-independent recognition model based on feature disentanglement and domain generalization. Zhao \emph{et al.}~\cite{zhao2023gan} proposed a novel two-stage supervised learning framework to calibrate the receiver-agnostic transmitter feature-extractor. Li \emph{et al.}~\cite{li2025receiver} addressed cross-receiver performance degradation through a two-stage unsupervised domain adaptation model and further enhancing robustness via multi-sample fusion inference adaptive few-sample selection strategy. Li \emph{et al.}~\cite{li2025agnostic} proposed a receiver-agnostic RF fingerprinting method using intra-domain and cross-domain prototypical contrastive learning to align the signals from different receivers. Zhou \emph{et al.}~\cite{zhou2025receiver} effectively extracted transmitter-related features using disentangled feature cross combination. 
 Feng \emph{et al.}~\cite{feng2025Cross} proposed a cross-receiver RFFI framework that dynamically aligns global and local feature distributions via Maximum Mean Discrepancy (MMD) over multi-scale features to alleviate receiver-induced domain shifts.

In the context of edge intelligence, where devices at the network edge possess limited computational resources, it is not feasible to retrain models from scratch on the edge nodes. Furthermore, bandwidth limitations and data privacy concerns exacerbate the challenge of transferring large amounts of labeled training data between edge nodes. To address these challenges, a more practical solution is to adapt the RFFI model to the new edge node with new receiver using only the pre-trained model and its own unlabeled data, without requiring access to labeled signals on the former edge node. This scenario, referred to as the Source-Data-Free Cross-Receiver RFFI (SCRFFI) problem, is highly relevant in edge intelligence settings, where computational and data transfer resources are constrained.
The SCRFFI literature remains very limited and has only emerged recently. Yang \emph{et al.}~\cite{yang2025cross} first formalized the SCRFFI setting and propose CSCNet, which adapts a pre-trained model to a new receiver using only unlabeled target signals via a three-branch loss combining entropy minimization, pseudo-label self-supervision, and contrastive learning. While Hu \emph{et al.}\cite{hu2025source} designed a source-free adaptation scheme based on multi-neighborhood semantic consistency and spatial adjacency. However, these works mainly emphasize local/instance-level consistency and do not explicitly handle class-prior shift, global prediction structure, or imbalanced/unknown target label distributions.

In this paper, we explore the SCRFFI problem within the context of edge intelligence and propose a novel approach to tackle it. Our theoretical analysis demonstrates that the performance of the adapted model on the new receiver is closely tied to the accuracy of the pseudo-labeling function. Furthermore, by continuously updating the pseudo-labeling function during adaptation, we show that the model's performance can be enhanced. Based on this insight, we introduce Momentum Soft pseudo-label Source Hypothesis Transfer (MS-SHOT), a method designed to refine the pseudo-labeling function in real-time. Our extensive experiments on real-world data demonstrate that MS-SHOT outperforms existing methods and offers a practical solution to the SCRFFI problem in edge intelligence applications. Our contributions are summarized as follows:

\begin{itemize}
\item We introduce the source-data-free cross-receiver RFFI problem—a practically important yet largely unexplored challenge in the RFFI literature. To bridge this gap, we propose a novel adaptation framework based on constrained pseudo-labeling and conduct a systematic investigation of SCRFFI, a setting that has received limited attention in prior research.
\item  We analyze the generalization capability of the SCRFFI problem and develop a theoretical error bound for the performance of the adapted model. We show that by judiciously refining the pseudo-labeling function during adaptation,  the adapted model can be improved in terms of generalization error probability on new receiver's data.  
\item Based on the theoretical analysis, we propose a novel method called MS-SHOT to timely update the pseudo-labeling function with a momentum-guided soft pseudo-labeling mechanism. We conducted extensive experiments on real-world data, and the results demonstrate that MS-SHOT outperforms existing methods. 
\end{itemize}

\subsection{Notations and Organization}
The $\mathbb{R}^{m\times n}$ denotes the set of matrices with dimension $m$-by-$n$, $\| \cdot\|_1$ and $\| \cdot \|_*$ denote the $\ell_1$ norm and the  nuclear norm, respectively, $\langle \cdot, \cdot \rangle$ represents the inner product, $[\bm a]_k$ denotes the $k$-th entry of the vector $\bm a$, $\mathds{1}_{a \neq b}$ is an indicator function with the value one if  $a\neq b$ is satisfied, and 0 otherwise, ${ \Delta}_K$ denotes the $(K-1)$-dimension  probability simplex, i.e. ${ \Delta}_K = \{ \bm z\in \mathbb{R}^K~|~ z_k \geq 0,  k=1,2,\cdots,K, \sum_{k=1}^K z_k = 1\}$, $\bm I_K\in \mathbb{R}^{K\times K}$, $\bm 1_K \in \mathbb{R}^K$ and $\bm e_k\in\mathbb{R}^K$ denote the identity matrix, the all-ones vector and the $k$-th unit vector, respectively, $\circledast$ denotes the convolution operator, $(\cdot)^T$ denotes the transpose operation, $\mathbb{E}[\,\cdot\,]$ denotes the expectation. The remainder of the paper is organized as follows. Section~\ref{sec:model} describes the signal model  and problem formulation; Section~\ref{sec:analysis} conducts the theoretical analysis of  generalization capability of the SCRFFI; Section~\ref{sec:method} describes the proposed MS-SHOT method; Section~\ref{sec:experiment} evaluates the performance of the MS-SHOT method with real-world data, and Section~\ref{sec:conclusion} concludes the paper.

\section{Signal Model Description and Problem Formulation} \label{sec:model}

Let us start from the  reception model. For simplicity, we consider single-carrier systems. Within the $\ell$-th time interval, the received RF signal, denoted as $x_\ell(t)$, can be described as
\begin{equation} \label{eq:system_model}
x_\ell(t) = \psi \left(c(t)  \circledast\varphi \left(s(t)\cos(\omega_0 t+\theta)\right)\right)+ n(t),
\end{equation}
where $(\ell-1)T \leq t\leq \ell T$, $n(t)$ is the noise, $T$ is the length of the interval, $\omega_0$ is the carrier frequency, $\theta$ is the initial phase, $s(t)$ is random modulating signal,  $c(t)$ denotes the channel response. Additionally, $\varphi(\cdot)$ models nonlinear distortion induced by the emitter's hardware, which essentially encodes the unique ``fingerprint'' of the emitter, while $\psi(\cdot)$ captures nonlinear reception characteristics resulting from the receiver's hardware. It should be noted that different receivers generally have distinct $\psi$.

To simplify our notation, we denote  $x_\ell^s$ as the $\ell$-th signal sample after pre-processing, and  $y_\ell^s\in {\cal K} \triangleq \{ 1,\ldots, K\}$ as the emitter index or label associated with $x_\ell^s$, where $K$ is the number of emitters and the superscript $s$  represents the source domain ${\cal D}^s$ corresponding to the pre-training receiver, denoted as Rx-1. Thus, the labeled training dataset for Rx-1 is succinctly represented as $\mathcal{S} = \{(x_1^s, y_1^s), \ldots, (x_{N^s}^s, y_{N^s}^s)\}$, where $N^s$ is the dataset size.
Similarly, the unlabeled dataset from the deployed receiver, referred to as Rx-2, is denoted as $\mathcal{T} = \{x_1^t, \ldots, x_{N^t}^t\}$, where  the superscript $t$ denotes the target domain ${\cal D}^t$ corresponding to Rx-2.

\begin{figure}[]
\centerline{\includegraphics[scale=0.5]{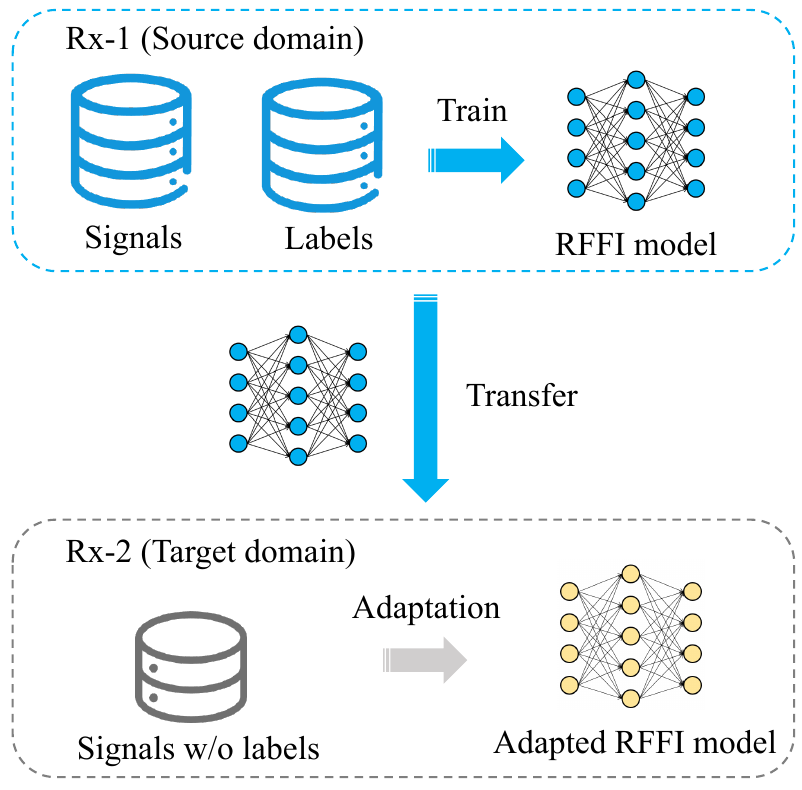}}
\vspace{-0.75em}
\caption{The scenario of source-data-free cross-receiver RFFI. Prior training commences in the source domain (Rx-1), where the RFFI model is developed using labeled signal data. Following model training in the source domain, a transfer phase occurs where only the learned model is applied to the target domain (Rx-2) due to data privacy. In the target domain, the model undergoes adaptation using signals without associated labels, resulting in an adapted RFFI model adapted to the new receiver.}
\label{fig:scen}
\vspace{-0.75em}
\end{figure}

The problem of source-data-free cross-receiver RFFI can be described as follows (see Fig.~\ref{fig:scen}):
Let $h^s: {\cal D}^s \rightarrow { \Delta}_K$ be the trained model on the labeled source domain ${\cal D}^s$. Given  $h^s$   and the unlabeled target data $\cal T$, we aim at finding an RFFI model $h^t: {\cal D}^t \rightarrow { \Delta}_K$ to minimize the classification error probability on the target domain ${\cal D}^t$, viz.,
\begin{equation} \label{eq:fstar}
\begin{aligned}
	& \min_{h^t \in {\cal H} } {\epsilon^t(h^t, f^t)} \triangleq {{\mathbb{E}}_{X^t \sim \mathcal{D}^t}\left[ \mathds{1}_{\arg\max {(h^t(X^t))} \neq \arg\max (f^t(X^t))}\right] } \\
	\approx &  \min_{h^t  \in {\cal H} }  {\hat\epsilon^t(h^t, f^t)}  \triangleq { \frac{1}{N^t}\sum_{x_i^t\in {\cal T}} \mathds{1}_{\arg\max (h^t(x_i^t)) \neq y_i^t}},
\end{aligned}
\end{equation}
where  $f^t: {\cal D}^t\rightarrow  \Delta_K$ denotes the (unknown) ground-truth label function on the target domain, i.e. $f^t(x^t) = \bm e_k$ if the signal $x^t$ is  from the emitter $k$, $\epsilon^t(h, f^t)$ and ${\hat{{\epsilon}}^t (h, f^t)}$ denote respectively the expected risk and the empirical risk on the target domain when a classification model $h$ is applied,  and ${\cal H}$ is a presumed hypothesis space consisting of the learning models from ${\cal D}^t$ to $\Delta_K$. We should mention that ${\cal H}$ is a functional space and its specific form depends on the structure and the size of the neural network. The  model's classification capability is determined by the complexity of  ${\cal H}$, usually measured by the Vapnik-Chervonenkis (VC) dimension~\cite{vapnik1998statistical}; we will elaborate on this in the next section.

%

The main difficulty of the problem in~\eqref{eq:fstar} lies in the unknown ground-truth label function $f^t$ or the target data label $y_i^t$, which makes it infeasible to use supervised learning method to find $h^t$. In the following, we will propose an indirect way to approximate $h^t$.


\section{Theoretical Analysis}\label{sec:analysis}

The problem~\eqref{eq:fstar} is not a well-posed problem because the true label function $f^t$ in the target domain is unavailable. Therefore, various surrogate formulations have been proposed in the literature to address this challenge. In this paper, we introduce a new formulation tailored to the problem, which is described as follows.

Since we do not have access to the true label function $f^t$ on the target domain, we instead utilize a pseudo-labeling function $\tilde{f}^t: \mathcal{D}^t \rightarrow \Delta_K$, derived from the source model $h^s$, to guide model adaptation. Our goal is to minimize the expected target-domain risk $\epsilon^t(h, f^t)$, but in practice, we minimize the empirical risk with respect to $\tilde{f}^t$, subject to certain regularization constraints to avoid overfitting to noisy pseudo-labels. In this section, we provide a theoretical analysis of this surrogate problem under two practical settings regarding the class distribution in the target domain.

We define the hypothesis output row-stochastic matrix $\bm Q_h \in \mathbb{R}^{N^t \times K}$ whose $i$-th row is $\bm Q_h(i,:) = h(x_i^t)$. The proposed SCRFFI formulation is described as follows:
\begin{subequations} \label{eq:main}
	\begin{align}
		\min_{h\in \mathcal{H}} & \quad \hat{\epsilon}^t(h, \tilde{f}^t) \\
		\text{s.t.} \quad & \| \bm Q_h \|_* \geq \zeta, \label{eq:main_b} \\
		& \|\bm{1}_{N^t}^T \bm Q_h - \bm{q}\|_1 \leq \gamma, \label{eq:main_c}
	\end{align}
\end{subequations}
where $\zeta$ is a rank-promoting lower bound on the nuclear norm, and $\gamma$ is a small scalar to allow tolerance in class proportion constraints.
Intuitively, the nuclear-norm constraint \eqref{eq:main_b} encourages $\bm Q_h$ to have a large nuclear norm and thus a high effective rank. Under the row-simplex constraint, this promotes rows that are close to one-hot vectors (confident, low-entropy predictions) while avoiding degenerate, low-rank solutions in which most samples collapse into a few dominant classes, thereby enforcing both confidence and diversity of the predictions. The $\ell_1$ constraint \eqref{eq:main_c} aligns the predicted class histogram $\bm{1}_{N^t}^T \bm Q_h$ with the target-domain prior $\bm q$ up to a small tolerance $\gamma$, which helps compensate for label shift and prevents the model from converging to a biased solution even when pseudo-labels are noisy.
The problem in~\eqref{eq:main} is a tractable surrogate problem for the problem in \eqref{eq:fstar}, and it minimizes the empirical risk (w.r.t. $\tilde{f}^t$) with some additional constraints. Since $\tilde{f}^t$ is in general not accurate, i.e. $\tilde{f}^t\neq {f}^t$, minimizing only $\hat{\epsilon}^t (h, \tilde{f}^t)$ is not sufficient to improve the model. As we will see in the following discussion, the constraints~\eqref{eq:main_b} and \eqref{eq:main_c} help learn a good model on the target domain.

The vector $\bm q$ in constraint~\eqref{eq:main_c} represents the target-domain class prior, i.e., the expected number of samples from each class in the target domain. For example, in a balanced setting where all $K$ classes have approximately equal samples, $\bm q$ is simply $[N^t/K, \ldots, N^t/K]^T$. This makes it straightforward to interpret $\bm q$ as describing the class distribution that the predictions of $h$ should approximately match at the global level.
In practice, the class distribution $\bm q$ may either be known beforehand or unknown during adaptation. In the following subsections, we analyze these two cases separately.

\subsection{Case 1: Class Distribution $\bm{q}$ in Target Domain is Known}
In this setting, we suppose that the class distribution in the target domain is known and represented as $\bm{q} = [\alpha_1, \alpha_2, \dots, \alpha_K] \cdot N^t$, where $\sum_k \alpha_k = 1$. 

Let $n_{k}=N^{t}\alpha_{k}$ be the expected sample count of class $k$.
With this prior information, the problem in \eqref{eq:main} can be rewritten as:
\begin{subequations} \label{eq:main_known}
	\begin{align}
		\min_{h\in \mathcal{H}} & \quad \hat{\epsilon}^t(h, \tilde{f}^t) \\
		\text{s.t.} \quad & \|\bm Q_{h}\|_{*}\;\ge\zeta(\bm{\alpha})\triangleq\sum_{k=1}^{K}\sqrt{n_{k}}, \label{eq:main_b_case1} \\
		& \bigl\|\bm 1_{N^{t}}^{\!\top}\bm Q_{h}-N^{t}\bm{\alpha}\bigr\|_{1}\le\gamma, \label{eq:main_c_case1}
	\end{align}
\end{subequations}
where $\zeta(\bm{\alpha})$ is a judiciously designed scalar depending on $\alpha$ and $N^t$. Constraint~\eqref{eq:main_c_case1} limits the deviation of the predicted class histogram from the ground-truth prior in the target domain. It should be noted that~\eqref{eq:main_b_case1} and~\eqref{eq:main_c_case1} play a key role in analyzing adaptation performance on the target domain. For simplicity, we consider the number of classes $K=2$ throughout this section. Then, we have the following generalization bound:
\begin{Theorem} \label{thm:1}
	Let  $\hat{h}$ be an optimal solution of the problem in ~\eqref{eq:main_known}. Suppose that the hypothesis space $\cal H$ has a VC dimension $d$. Then, for any $\rho\in (0,~1)$, with probability at least $1-\rho$, the following inequality holds 
	\begin{equation} \label{eq:Thm1}
		\epsilon^t(\hat{h}, f^t) \leq 2\epsilon^t(\tilde{f}^t, f^t) + c_1  
	\end{equation}
	where $c_1 = 2\sqrt{\frac{d(\log(2N^t/d)+1) + \log(4/\rho)}{N^t}}$.
\end{Theorem}

The main idea of the proof of Theorem~\ref{thm:1} is as follows. We first show that the optimization problem is well-posed by verifying that the (unknown) ground-truth labeling function $f^t$ is a feasible solution. Then, we use the triangle inequality to break down the target-domain risk into the pseudo-labeling error and the empirical risk minimized during optimization. Finally, we apply a standard VC-dimension-based generalization bound~\cite{vapnik1998statistical, ben2010theory} to connect the empirical risk with the expected risk. The full proof is provided in the supplementary file.
In words,  the inequality in~\eqref{eq:Thm1} reveals that the expected classification error probability of $\hat{h}$ on the target domain ${\cal D}^t$ is upper bounded by some constant, which is related to the number of data samples $N^t$, the capacity $d$ of the hypothesis $\cal H$ and  the  accuracy of the pseudo-label function $\tilde{f}^t$. In particular,  $c_1$ is generally a fixed number, independent of the learning method. Therefore, to minimize the upper bound, it suffices to make the pseudo-label function $\tilde{f}^t$ as accurate as possible. 
If $\tilde{f}^t = f^t$, then $\epsilon^t(\tilde{f}^t, f^t)$ diminishes.

Let us consider a special case $\tilde{f}^t = h^s$, i.e. using the source model to perform the adaptation. We have the following result.
\begin{Theorem}\label{thm:2}
	Under the same setting in Theorem~\ref{thm:1}, for any $\rho\in (0,~1)$, with probability at least $1-\rho$, the following inequality holds 
	\begin{equation} \label{eq:upper_bound_hs}
		\epsilon^t(\hat{h}, f^t) \leq 2 \epsilon^s({h}^s, f^s) +  c_2 
	\end{equation}
	where $c_2 =  2\eta^\star + d_{\cal H}({\cal D}^s, {\cal D}^t) + c_1$,  $\eta^\star = \min_{h \in \mathcal{H}}{\epsilon^s(h, f^s) + \epsilon^t(h, f^t)}$ and $d_{\mathcal{H}}({\cal D}^s, {\cal D}^t) = 2\sup_{h,h' \in \mathcal{H}}{ | {\epsilon}^t(h, h') - {\epsilon}^s(h, h')}|$.
\end{Theorem}

The proof relates the target risk to the source risk using the triangle inequality and domain discrepancy, along with a VC-dimension-based generalization bound~\cite{ben2010theory}. The full details are provided in the supplementary file.
We should mention that the right-hand side of~\eqref{eq:upper_bound_hs} is a constant. The $\epsilon^s({h}^s, f^s) $ corresponds to the expected classification error probability of the source model  on the source domain, which is unrelated to the target domain. The constant $c_2$ is related to $\eta^\star$ and  $d_{\cal H}({\cal D}^s, {\cal D}^t)$. In particular,   $\eta^\star$ represents the combined expected classification error probability when the model is trained jointly with the labeled source data and the labeled target  data. The $d_{\cal H}({\cal D}^s, {\cal D}^t)$ measures the domain discrepancy of ${\cal D}^s$ and ${\cal D}^t$ with respect to  $\mathcal{H}$. When ${\cal D}^s$ is close to ${\cal D}^t$, $d_{\cal H}({\cal D}^s, {\cal D}^t)$ approaches zero, which implies a smaller $\epsilon^t(\hat{h}, f^t) $ can be attained, i.e. the learned model $\hat{h}$  can better adapt to the target domain.

Combining Theorems~\ref{thm:1} and \ref{thm:2}, we immediately have the following corollary.
\begin{Corollary} \label{corrollay}
	Given a pseudo-label function $\tilde{f}^t$ satisfying $\epsilon^t(\tilde{f}^t, f^t)< \epsilon^t(h^s, f^t)$, it holds that 
	\begin{equation} \label{eq:corrollay}
		\epsilon^t(\hat{h}, f^t) < 2 \epsilon^s({h}^s, f^s) + c_2.
	\end{equation}
	That is, if $\tilde{f}^t$ is more accurate than $h^s$,   the resultant expected classification error probability on the target domain can be improved.
\end{Corollary}

Corollary~\ref{corrollay} is a direct consequence of Theorems~\ref{thm:1} and~\ref{thm:2}. The detailed proof can be  found in the supplementary file.  Thus far, we have analyzed the target-domain performance of  $\hat{h}$ learned from the problem in \eqref{eq:main}. Concisely speaking, $\hat{h}$ has a provable performance guarantee on the target domain, and the performance depends on several factors. Among them, the accuracy of the pseudo-labeling function $\tilde{f}^t$ is most important. It is plausible to choose the  source model $h^s$ as the initial  $\tilde{f}^t$ and gradually improve  $\tilde{f}^t$ to attain a better adapted model.

\subsection{Case 2: Class Distribution $\bm{q}$ in Target Domain is Unknown}

When the true class prior of the target domain is unavailable, 
we propose to estimate the $\bm q$ with the pseudo-labeling function $\tilde{f}^t$ as:
\[
\hat{\bm q}
\;=\;
\bm 1_{N^{t}}^{\!\top}\bm Q_{\tilde f^{t}}
=[\hat n_{1},\ldots,\hat n_{K}].
\]
and modify the problem~\eqref{eq:main} as
\begin{subequations} \label{eq:main_unknown}
	\begin{align}
		\min_{h\in \mathcal{H}} & \quad \hat{\epsilon}^t(h, \tilde{f}^t) \\
		\text{s.t.} \quad & \|\bm Q_{h}\|_{*}\ge\zeta(\hat{\bm q})\triangleq\sum_{k=1}^{K}\left(\sqrt{\hat n_{k}} - \dfrac{\gamma}{2\sqrt{\hat n_{\min}}}\right), \label{eq:main_b_case2} \\
		& \bigl\|\bm 1_{N^{t}}^{\!\top}\bm Q_{h}-\hat{\bm q}\bigr\|_{1}\le \gamma, \label{eq:main_c_case2}
	\end{align}
\end{subequations}
with $\hat n_{k}$ is the number of samples classified by $\tilde{f}^t$ as class $k$, and $\hat n_{\min}=\min_{k}\hat n_{k}$.

To analyze problem~\eqref{eq:main_unknown}, we draw the following assumption:
\begin{Assum}\label{assum:distribution}
	The class distribution predicted by the pseudo-labeling function $\tilde{f}^t$  is $\gamma$-close to the true class distribution in the target domain:
	\begin{equation}
		\bigl\|\bm 1_{N^{t}}^{\!\top}\bm Q_{f^{t}}
		-\hat{\bm q}\bigr\|_{1}\le\gamma
	\end{equation}
	and $\gamma<2\hat n_{\min}$.
\end{Assum}

Then, we have the following theorem:
\begin{Theorem} \label{thm:unknown}
	Under Assumption~\ref{assum:distribution}, let $\hat{h}$ be an optimal solution of problem~\eqref{eq:main_unknown}. Then, the same bounds in \eqref{eq:Thm1}, \eqref{eq:upper_bound_hs} and \eqref{eq:corrollay} still hold.
\end{Theorem}
%
%

The proof of Theorem~\ref{thm:unknown} is essentially the same as Theorems~\ref{thm:1}, \ref{thm:2} and Corollary~\ref{corrollay}. The detailed proof is given in the supplementary file.

In both the known and unknown class prior settings, our theoretical analysis arrives at the same key result. Regardless of whether the true class distribution is known or approximated, the constraints in our optimization formulation, which promote confident and diverse predictions while aligning with global class proportions, ensure that if $\tilde{f}^t$ is more accurate than the source model $h^s$, then the learned model $\hat{h}$ will achieve improved performance on the target domain. This result not only justifies the role of pseudo-label refinement, but also highlights the importance of incorporating global structural regularization to mitigate the adverse effects of label noise during adaptation.

\begin{figure*}[htbp]
	\centerline{\includegraphics[scale=0.43]{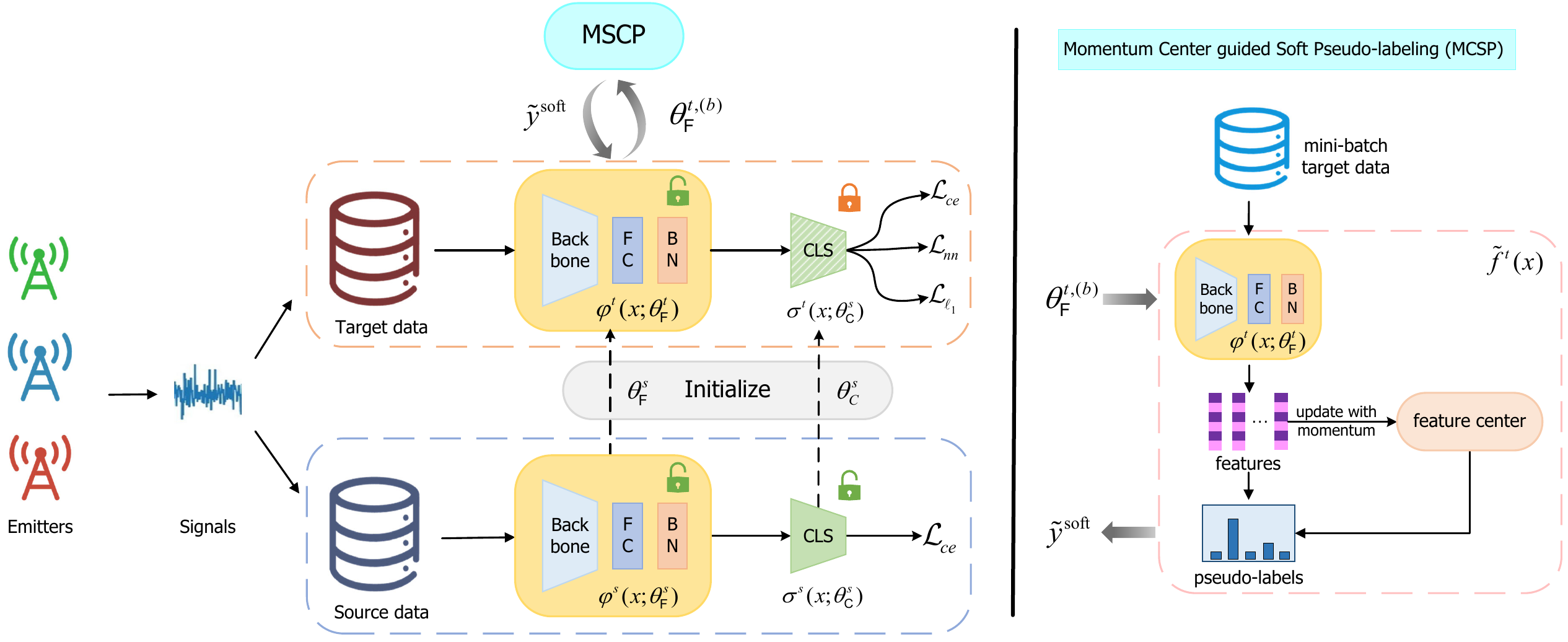}}
	\caption{The overview of MS-SHOT. The parameters of the adapted model $h^t$ are first initialized with the feature extractor parameters $\theta_{\sf F}^s$ and classifier parameters $\theta_{\sf C}^s$ of the model trained in the source domain. Following this, the classifier parameters in $h^t$ are frozen to preserve the decision boundary, while updates are confined to the feature extractor parameters. Throughout the adaptation phase, the model interacts with Momentum Center guided Soft Pseudo-labeling (MCSP) during each batch to procure real-time soft pseudo-labels $\tilde{y}^{\rm soft}$. Loss functions $\mathcal{L}_{ce}$, $\mathcal{L}_{nn}$, and $\mathcal{L}_{\ell_1}$ are then employed to calculate the loss, and gradient descent is utilized to iteratively update $\theta_{\sf F}^t$ until the model achieves convergence (FC: Fully-Connected layer, BN: BatchNorm, CLS: Classifier).}
	\label{fig:overview}
\end{figure*}

\section{Proposed Method}\label{sec:method}

\subsection{Overview of the proposed method}

The overview of the proposed method is shown in Fig.~\ref{fig:overview}, where the  middle part illustrates the model structure (upper: source model $h^s$, lower: target model $h^t$). The source model $h^s$ takes the following form:
\begin{equation} \label{eq:hs}
h^s(x; \theta_{\sf F}^s,\theta_{\sf C}^s ) =   \sigma^s \big(  \varphi^s(x;  \theta_{\sf F}^s); \theta_{\sf C}^s \big)
\end{equation} 
where $\varphi^s :  {\cal D}^s \rightarrow {\mathbb{R}}^d$  and $\sigma^s: {\mathbb{R}}^d \rightarrow \Delta_K$ denote the feature extractor (with  parameter $\theta_{\sf F}^s$) and the classifier (with  parameter $\theta_{\sf C}^s$), respectively. We should mention that the feature extractor transforms raw RF signals into a high-dimensional feature space, accentuating the intrinsic characteristics of the emitters. The feature extractor can be implemented by commonly used generic deep neural networks, e.g. CNN or ResNet. The classifier converts the feature vector into a probability vector, rendering the classification confidence for the current input signal.   The classifier is typically realized with a multi-layer perceptron followed by Softmax operation. The implementation details on $\varphi^s$ and $\sigma^s$ will be given in the experimental section. 

Similarly, the adapted model $h^t$ on the target domain is represented as
\begin{equation} \label{eq:ht}
h^t(x; \theta_{\sf F}^t,\theta_{\sf C}^t )  = \sigma^t \big(\varphi^t(x;  \theta_{\sf F}^t); \theta_{\sf C}^t \big).
\end{equation} 
The proposed adaptation procedure is summarized as follows.
\begin{itemize}
\item {\bf Step 1: } {Input the source model $h^s(\cdot; \theta_{\sf F}^s, \theta_{\sf C}^s)$ and $\cal T$;}
\item {\bf Step 2: } Copy the source model $h^s$ to the target model $h^t$ and initialize $h^t \leftarrow h^s$;
\item {\bf Step 3: } Set the pseudo-label function $\tilde{f}^t \leftarrow h^s$; 
\item {\bf Step 4: } Freeze the target network's  classifier $\theta_{\sf C}^t= \theta_{\sf C}^s$;
\item {\bf Step 5: } Update only the feature extractor $\theta_{\sf F}^t$ according to~\eqref{eq:main};
\item {\bf Step 6: } Refine the pseudo-label function $\tilde{f}^t$ based on the updated  feature extractor $\theta_{\sf F}^t$;
\item {\bf Step 7: } Repeat Steps 5--6 until convergence;
\item {\bf Step 8: } Output the target model $h^t(\theta_{\sf F}^t, \theta_{\sf C}^t)$.
\end{itemize}

As seen, Steps 5 and 6 are the key to our approach and we will elaborate on them in the following subsections.

\begin{rmk}
In Step 4, we follow~\cite{liang2020we}, where the classifier of the target network is frozen, and its parameters remain unchanged during adaptation. Only the parameters of the feature extractor of the target network are updated. This strategy is intended to stabilize the classification boundary, thereby aligning the distribution of signal features between the source and target domains.
\end{rmk}

\subsection{Update of the feature extractor $\theta_{\sf F}^t$}

The update of the feature extractor $\theta_{\sf F}^t$ is based on the constrained optimization problem introduced in Section~\ref{sec:analysis}, which aims to minimize the empirical risk under pseudo-label supervision while ensuring consistency with the overall label distribution and avoiding trivial solutions. Depending on whether the class prior in the target domain is known or estimated, we formulate a unified optimization problem with different constraint forms. For clarity, we restate the general form of the problem:
\begin{subequations} \label{eq:feature-update}
	\begin{align}
		\min_{ \theta_{\sf F}^t} & ~ \hat{\epsilon}^t (\theta_{\sf F}^t, \tilde{f}^t) \label{eq:feature-update-a}\\
		{\rm s.t. } & ~ \| \bm Q_{\theta_{\sf F}^t} \|_* \geq \zeta, \label{eq:feature-update-b} \\
		& ~ \left\| \bm 1_{N^t}^T \bm Q_{\theta_{\sf F}^t} - \bm{q} \right\|_1 \le \gamma, \label{eq:feature-update-c}
	\end{align}
\end{subequations}
where for notational convenience  we have used $\theta_{\sf F}^t$ to indicate the target model $h^t$, and $\bm{q}$ denotes a reference class distribution vector that depends on the setting:
\begin{itemize}
	\item If the target-domain class prior $\bm \alpha = [\alpha_1, \dots, \alpha_K]$ is known, then $\bm{q} = N^t \bm \alpha$;
	\item If the class prior is unknown, then $\bm{q} = \hat{\bm q} = \bm 1_{N^t}^{\top} \bm Q_{\tilde{f}^t}$ is the estimated class count vector based on pseudo-labels.
\end{itemize}

The problem in \eqref{eq:feature-update} is a constrained optimization problem with  non-smooth objective $\hat{\epsilon}^t (\theta_{\sf F}^t, \tilde{f}^t)$, which is not convenient to optimize for neural networks. To circumvent this difficulty, we consider the following unconstrained smoothed approximation of the problem in \eqref{eq:feature-update}, viz.,
\begin{equation} \label{eq:total}
	\min_{\theta_{\sf F}^t} ~\mathcal{L}(\theta_{\sf F}^t) =  \lambda_1 \mathcal{L}_{ce} + \lambda_2 \mathcal{L}_{nn} + \lambda_3 \mathcal{L}_{\ell_1},
\end{equation}
with $\lambda_i>0~ (i=1,2,3)$ being  hyper-parameters and
\begin{subequations}\label{eq:def_loss}
\begin{align}
	\mathcal{L}_{ce} & \triangleq  - \frac{1}{N^t}\sum_{x \in{\cal T}} \left[  \sum_{k=1}^K [\tilde{f}^t(x)]_k \log \left({ [h^t(x; \theta_{\sf F}^t )]_k}\right) \right], \label{eq:def_loss_ce}  \\
	\mathcal{L}_{nn} & \triangleq - \| \bm Q_{\theta_{\sf F}^t} \|_*, \label{eq:def_loss_nn} \\
	\mathcal{L}_{\ell_1} & \triangleq \left\| \bm 1_{N^t}^T \bm Q_{\theta_{\sf F}^t} - \bm q \right\|_{1}, \label{eq:def_loss_l1}
\end{align}
\end{subequations}
where  $\mathcal{L}_{ce}$ is the cross-entropy between $\tilde{f}^t(x)$ and $ h^t(x; \theta_{\sf F}^t )$, which is typically used for classification tasks; $	\mathcal{L}_{nn}$ and $\mathcal{L}_{\ell_1}$ are penalties corresponding to the constraints~\eqref{eq:feature-update-b}  and \eqref{eq:feature-update-c}, respectively.

The problem in \eqref{eq:total} is typically handled by stochastic gradient descent (SGD) method, e.g. Adam~\cite{kingma2014adam}:
\[  \theta_{\sf F}^{t, (b+1)} \leftarrow \theta_{\sf F}^{t, (b)} - \eta^{(b)} \nabla_{\theta_{\sf F}^t} \mathcal{L}(\theta_{\sf F}^{t, (b)})  \]
where $b$ denotes the $b$-th mini-batch, $\nabla_{\theta_{\sf F}^t} \mathcal{L}(\theta_{\sf F}^{t, (b)})$ denotes the stochastic gradient evaluated within the $b$-th mini-batch data   ${\cal T}_b$ ($\cup_{b}{\cal T}_b = {\cal T} $), and  $ \eta^{(b)}>0$ is the step-size.

\subsection{Update of the pseudo-label $\tilde{f}^t$}
As mentioned before, the accuracy of the pseudo-label $\tilde{f}^t$ is crucial for adaptation. After updating $\theta_{\sf F}^t$ for each mini-batch, a timely update of the pseudo-label $\tilde{f}^t$ is necessary. To this end, we propose a novel MS-SHOT method to update  $\tilde{f}^t$. The MS-SHOT is inspired by Source HypOthesis Transfer (SHOT)~\cite{liang2020we} method  with additional Momentum Center-Guided Soft Pseudo-labeling idea in order to achieve more timely and flexible update of $\tilde{f}^t$ than the original SHOT.

To better elaborate on MS-SHOT, let us briefly introduce the SHOT method. The SHOT method employs  k-means to update  the pseudo-label of the target signals for every training epoch. Specifically,  at the start of every training epoch, SHOT sets the pseudo-label model $\tilde{f}^t$   to the latest trained target model $h^t$, and   computes the initial pseudo-label of the target  signals via cosine similarity:
\begin{equation} \label{eq:y_hat}
\tilde{y}^{\rm init} (x) = \mathop{\arg\max}_{k\in \{1,\ldots, K\}} {\frac{\langle  \varphi^t(x), c_k^{\rm init} \rangle}{\| \varphi^t(x)\|\|c_k^{\rm init}\|}}, ~~\forall~x\in {\cal T},
\end{equation}
where $c_k^{\rm init} \in \mathbb{R}^d$ is the initial  center of class $k$, given by
\begin{equation} \label{eq:c_init}
c_k^{\rm init} = \frac{\sum_{x \in \mathcal{T}} { [\tilde{f}^t(x)]_k \varphi^t(x)}} {\sum_{x \in \mathcal{T}}  [\tilde{f}^t(x)]_k}, \quad k=1,\ldots,K.
\end{equation}
Then, the  pseudo-labels of the target signals are obtained by k-means approach:
\begin{equation} \label{eq:SHOT}
[\tilde{y}(x)]_k  = \begin{cases}
	1, & {\rm if}~k = k^\star,\\
	0, & {\rm otherwise},
\end{cases}
\end{equation}
where 
\begin{equation} \label{eq:c_k}
\begin{aligned} 
	k^\star & =   \mathop{\arg\max}_{k\in \{1,\ldots, K\}} {\frac{\langle \varphi^t(x), c_k \rangle}{\| \varphi^t(x)\|\|c_k\|}},\\
	c_k & = \frac{\sum_{x \in \mathcal{T}} {\mathds{1}_{ \tilde{y}^{\rm init}(x) = k} \cdot \varphi^t(x)}} {\sum_{x \in \mathcal{T}} {\mathds{1}_{ \tilde{y}^{\rm init}(x) = k}}},
\end{aligned} 
\end{equation}
for all $x\in {\cal T}$. The pseudo-label $\tilde{y}(x)$ is used   for evaluating the cross-entropy loss ${\cal L}_{ce}$ in~\eqref{eq:def_loss_ce} and guiding the update  of the target model $h^{t}$ in the next epoch.  Since SHOT makes hard-decision on the label $\tilde{y}(x)$, it usually suffers from error propagation, if pseudo-labels are incorrect, especially at the early training stage.  Furthermore, SHOT  updates the pseudo-labels  at the epoch level, and the incorrect pseudo-labels may have a long-lasting impact on the subsequent training. In view of the above drawbacks, we propose Momentum Center guided Soft Pseudo-labeling strategy to reduce the impact of label noise and to update pseudo-labels timely.

\begin{algorithm}[t]
	\caption{ The MS-SHOT for SCRFFI}
	\label{method}
	\textbf{Input:}
	\begin{algorithmic}
		\State $\mathcal{T}$: Unlabeled target domain data $\{{(x_i^t)}\}_{i=1}^{N^t}$
		\State $h^s$: The model trained on the source domain
		\State $\eta$: Learning rate
		\State $\lambda_1, \lambda_2, \lambda_3, \tau, \beta$: Hyper-parameters
	\end{algorithmic}
	\textbf{Output:} Parameters $\theta_{\sf F}^t, \theta_{\sf C}^t$ of the target model $h^t$
	\begin{algorithmic}[1]
		\State Initialize $\theta_{\sf F}^t, \theta_{\sf C}^t$ with the source model $h^s$
		\State Initialize the pseudo-label function $\tilde{f}^t$ using $h^s$
		\State Freeze the target model's classifier $\theta_{\sf C}^t = \theta_{\sf C}^s$
		\For {$epoch = 1,2,\cdots$}
		\State  Compute feature center $c_k$ by \eqref{eq:c_init}, \eqref{eq:c_k}
		\State  Get class distribution $\bm q$ or its estimate $\hat{\bm q}$
		\For {$b = 1, 2, \cdots$}
		\State Get a batch of data ${\cal T}_b$ from $\mathcal{T}$
		\State Update feature center $c_k^{(b)}$ by \eqref{eq:update_ck}
		\State Update  $\tilde{y}^{\rm soft}(x)$ for $x\in {\cal T}_b$ by \eqref{eq:soft_pseudo}
		\State Calculate the loss $\mathcal{L}(\theta_{\sf F}^t) =  \lambda_1 \mathcal{L}_{ce} + \lambda_2 \mathcal{L}_{nn} + \lambda_3 \mathcal{L}_{\ell_1}$
		\State $\theta_{\sf F}^{t, (b)} \leftarrow \theta_{\sf F}^{t, (b-1)} - \eta^{(b)} \nabla_{\theta_{\sf F}^t} \mathcal{L}(\theta_{\sf F}^{t, (b-1)})$
		\State Update $\tilde{f}^t \gets h^t(\theta_{\sf F}^{t, (b)}, \theta_{\sf C}^s)$
		\EndFor
		\EndFor
		\State \Return $\theta_{\sf F}^t, \theta_{\sf C}^t$
	\end{algorithmic}
\end{algorithm}

In MCSP, the pseudo-labels are updated in soft-decision and mini-batch manners. Given the $b$-th mini-batch of the target  signals ${\cal T}_b$, the feature center $c_k$ is updated per batch with momentum as follows:
\begin{equation} \label{eq:update_ck}
c_k^{(b)} \leftarrow \beta c_k^{(b-1)} + (1 - \beta) \frac{\sum_{x \in \mathcal{T}_b} { [\tilde{f}^t(x)]_k\varphi^t(x)}} {\sum_{x \in \mathcal{T}_b} { [\tilde{f}^t(x)]_k}},
\end{equation}
where $\beta \in [0, 1)$ is a momentum coefficient, $c_k^{(b-1)}$ is feature center after the $(b-1)$-th mini-batch update. Then the soft pseudo-label $\tilde{y}^{\rm soft} \in \Delta_K$ is calculated as
\begin{equation} \label{eq:soft_pseudo}
[\tilde{y}^{\rm soft}(x)]_k = \frac{e^{ \frac{\langle \varphi^t(x), c_k^{(b)} \rangle}{\tau \| \varphi^t(x)\|\|c_k^{(b)}\|}}}{\sum_{i=1}^K e^{ \frac{\langle \varphi^t(x), c_i^{(b)} \rangle}{\tau \| \varphi^t(x)\|\|c_i^{(b)}\|}}}, ~~k=1,\ldots, K,
\end{equation}
for all $x\in {\cal T}_b$,	where $\tau > 0$ is a temperature parameter.

To sum up, the complete description of MS-SHOT is shown in Algorithm~\ref{method}.

\begin{figure} \centering
	\subfloat[]{\includegraphics[width=0.7\columnwidth]{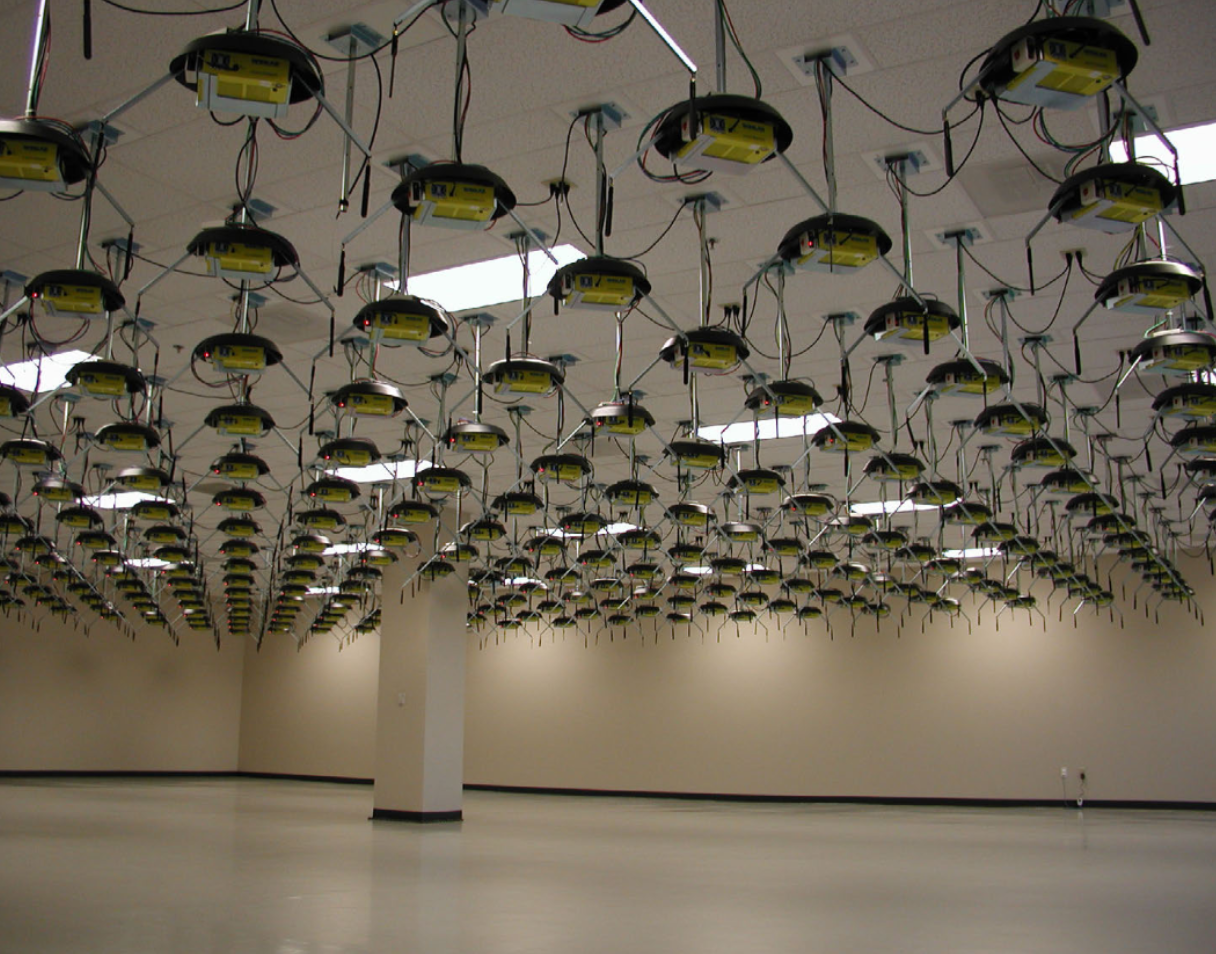}
		\label{fig:wisig_photo}} \quad
	\subfloat[]{\includegraphics[width=0.45\columnwidth]{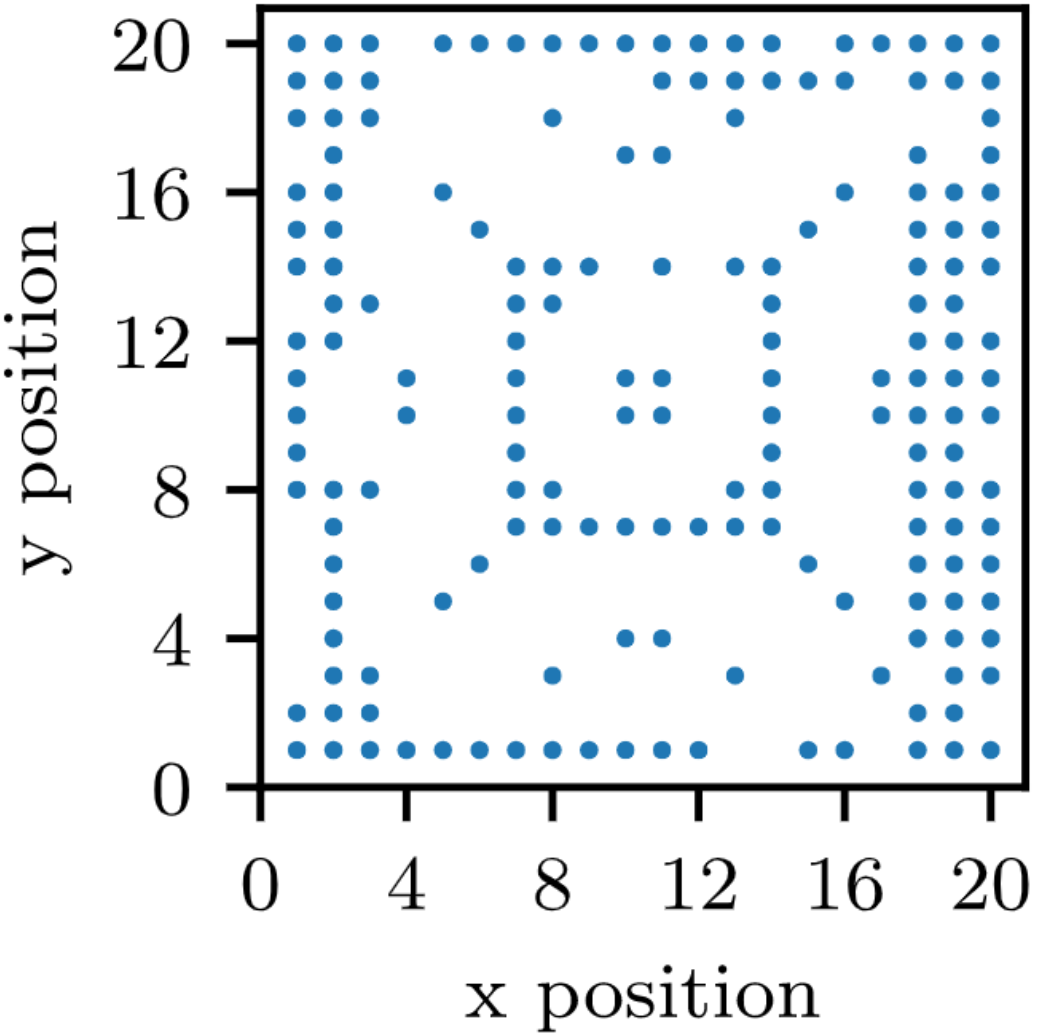}
		\label{fig:wisig_tx}}
	\subfloat[]{\includegraphics[width=0.45\columnwidth]{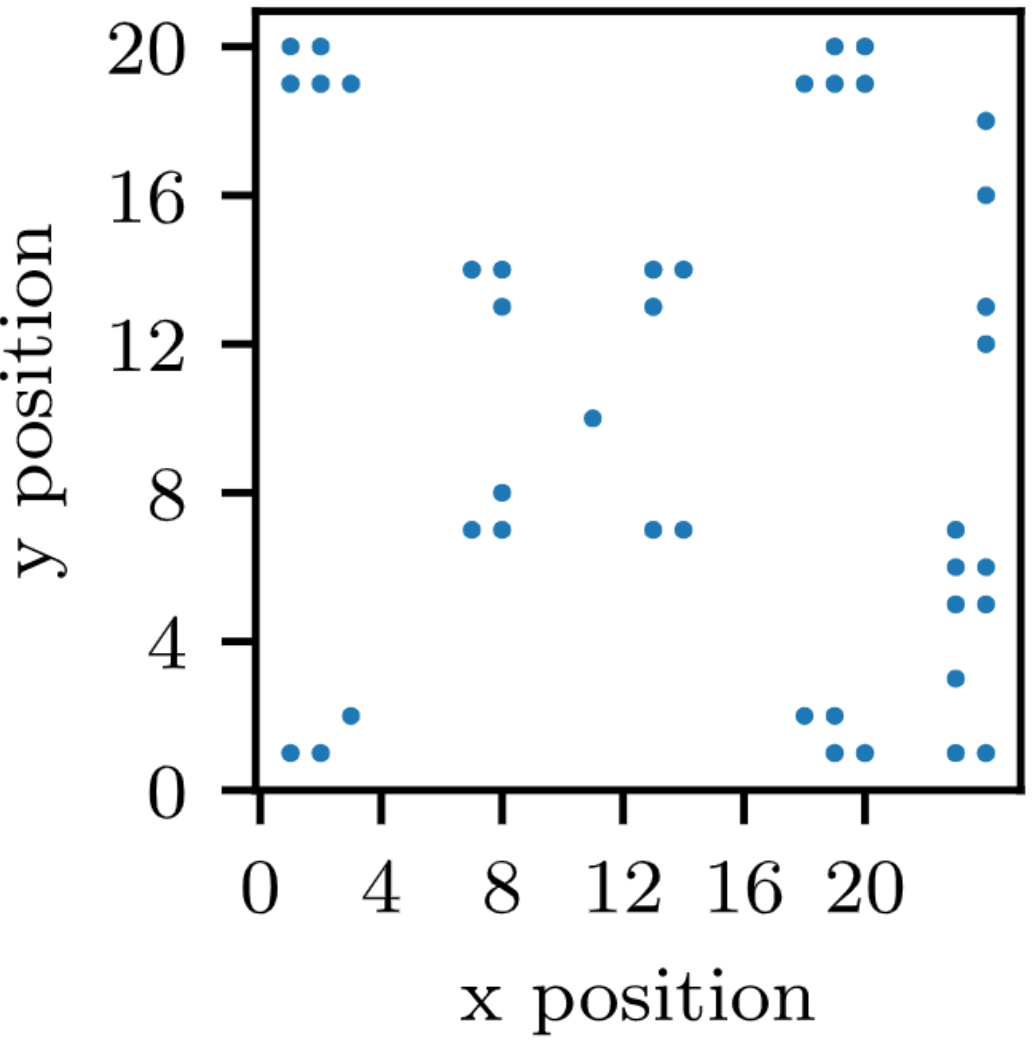}
		\label{fig:wisig_rx}}
	\caption{Scenario description of Wisig dataset~\cite{hanna2022wisig}. (a) Nodes are arranged as a grid. Each node is a roof-mounted PC with at least one WiFi emitter (Tx). Some nodes are equipped with USRP receivers (Rx). (b) Positions of Tx. (c) Positions of Rx.}
	\label{fig:wisig}
\end{figure}

\section{Experiment} \label{sec:experiment}

\begin{table}[t]
	\renewcommand\arraystretch{1}
	\centering
	\caption{\label{setup}The setup in the experiments}
	\begin{tabular}{c c c}
		\toprule
		\textbf{Item} & \textbf{Hyper-parameters} & \textbf{Values}  \\
		\hline
		Wisig Dataset & signal size & $2 \times 256$ \\
		\hline
		HackRF Dataset & signal size & $2 \times 28,000$ \\
		\hline
		\multirow{5}{*}{Feature extractor} & Layers & 18 \\
		\cline{2-3}
		~ & Kernel size & 3 \\
		\cline{2-3}
		~ & Output size & 1024 \\
		\cline{2-3}
		~ & Number of parameters & 2.1M\\
		\cline{2-3}
		~ & FLOPs & 130.747M\\
		\hline
		\multirow{4}{*}{Training} & Optimizer & Adam \\
		\cline{2-3}
		~ & Epoch & 20 \\
		\cline{2-3}
		~ & Batch size & 64 \\
		\cline{2-3}
		~ & Learning rate & 0.0006 \\
		\bottomrule
	\end{tabular}
\end{table}

\subsection{Experimental Setup\protect\footnote{The source code is available at https://github.com/YannLeo/MS-SHOT.}}

\subsubsection{Dataset}

This paper uses two datasets, namely the Wisig dataset~\cite{hanna2022wisig}\footnote{https://cores.ee.ucla.edu/downloads/datasets/wisig/} and the HackRF dataset created by ourselves, to evaluate the performance.

The Wisig dataset comprises 10 million packets captured from 174 off-the-shelf WiFi emitters (Tx) and 41 USRP receivers (Rx) over the course of one month. Fig.~\ref{fig:wisig} shows the scenario description of Wisig. Tx and Rx are deployed on the nodes, which are arranged as a grid. And the 2D-coordinates scattergram of Tx and Rx are shown in Fig.~\ref{fig:wisig}\subref{fig:wisig_tx} and Fig.~\ref{fig:wisig}\subref{fig:wisig_rx}. The signals are obtained by deploying Tx and Rx on nodes configured in a grid pattern. We focus on ManySig, a specific subset of the Wisig dataset consisting of 1,000 equalized signals from all Tx-Rx pairs, including 6 Tx and 12 Rx, over four days. Each signal is comprised of both in-phase (I) and quadrature-phase (Q) channels with 256 discrete sampling points. Our Wisig dataset experiments comprise two types of tests. The first is the cross-receiver test, where 12 domains corresponding to each receiver are used to capture signals over four days. The second is the cross-day test, where the four domains represent each capture day with all 12 Rx. The purpose of these two tests is to evaluate the performance of MS-SHOT in the presence of data distribution bias. 
To uniquely represent Tx and Rx, they can be assigned two-dimensional (2D) coordinates since they are situated on a 2D grid. 
The notation ``$x-y$'' denotes a receiver physically located at coordinates $(x, y)$ in the Wisig deployment grid. In each domain adaptation task, ``$x-y \rightarrow x'-y'$'' indicates that the model is adapted from data collected at receiver $(x, y)$ (source domain) to data collected at receiver $(x', y')$ (target domain). To ensure robustness and reflect practical deployment scenarios, we use signal samples recorded over four consecutive days for both the source and target receivers. This multi-day setup mitigates potential temporal biases and enhances generalization across different collection conditions.

The HackRF, a cost-effective open-source software radio platform, was utilized to generate and collect the HackRF dataset, which comprises signal data. Data collection involved 4 HackRFs acting as transmitting IoT devices and 3 HackRFs (named h1, h2, and h3) serving as receivers, all classified as IoT devices. Signals were generated using MATLAB and modulated using BFSK with a sampling rate of 2 MHz. To introduce randomness, amplitude and frequency noise were uniformly added to the signals.
Subsequently, the transmitted signals traversed a relatively stable channel before being converted to Intermediate Frequency (IF) for reception. The collected signals were then transformed into time-domain samples using data frames, with each sample containing 28,000 sampling points. In the experiments, the notation ``h1 $\rightarrow$ h2'' represents the adaptation from signals collected at receiver ``h1'' to that at ``h2''.

\begin{table*}[htbp]
	\renewcommand\arraystretch{1}
	\centering
	\caption{\label{results_other_methods}Classification accuracies (\%) comparison with existing methods on the Wisig dataset}
	\begin{tabular}{l c c c c c c c}
		\toprule
		Method & Source-data-free & 14-7 $\rightarrow$ 3-19 & 1-1 $\rightarrow$ 1-19 & 2-1 $\rightarrow$ 18-2 & 7-7 $\rightarrow$ 8-8 & 2-19 $\rightarrow$ 19-2 & 20-1 $\rightarrow$ 7-14\\
		\midrule
		Source only & & 33.26 $\pm$ 7.12 & 59.86 $\pm$ 5.03 & 21.78 $\pm$ 2.64 & 53.53 $\pm$ 1.70 & 62.91 $\pm$ 4.98 & 72.47 $\pm$ 6.13\\
		DANN \cite{ganin2016domain} & & 62.39 $\pm$ 1.98 & 77.77 $\pm$ 0.19 & 47.60 $\pm$ 6.92 & 79.46 $\pm$ 1.23 & 83.36 $\pm$ 1.34 & 98.37 $\pm$ 0.04\\
		MCD \cite{saito2018maximum} & & 66.24 $\pm$ 0.10 & 79.64 $\pm$ 0.03 & 48.62 $\pm$ 0.04 & 66.95 $\pm$ 0.14 & 78.13 $\pm$ 0.03 & 82.39 $\pm$ 0.03 \\
		SHOT \cite{liang2020we} & \checkmark & 76.99 $\pm$ 0.03 & 78.69 $\pm$ 0.00 & 45.02 $\pm$ 0.05 & 95.75 $\pm$ 0.00 & 89.97 $\pm$ 0.03 & 99.43 $\pm$ 0.00\\
		CoNMix \cite{kumar2023conmix} & \checkmark & \textbf{80.74 $\pm$ 0.15} & 66.70 $\pm$ 0.00 & 43.58 $\pm$ 0.00 & 66.51 $\pm$ 0.00 & 68.59 $\pm$ 1.09 & \underline{99.64 $\pm$ 0.02}\\
		GPUE \cite{litrico2023guiding} & \checkmark & 72.42 $\pm$ 0.37 & 65.53 $\pm$ 0.13 & 49.71 $\pm$ 0.66 & 81.71 $\pm$ 0.21 & 76.60 $\pm$ 0.28 & 96.07 $\pm$ 0.17\\
		UPA \cite{chen2024uncertainty} & \checkmark	& 63.28 $\pm$ 0.12 & 73.29 $\pm$ 0.02 & \underline{50.99 $\pm$ 0.07} & 72.01 $\pm$ 0.10 & 84.05 $\pm$ 0.13 & 97.19 $\pm$ 0.04\\
		SFADA \cite{HE2024110246} & \checkmark & 42.74 $\pm$ 0.09 & 74.83 $\pm$ 0.15 & 20.81 $\pm$ 0.36 & 73.15 $\pm$ 0.25 & 82.04 $\pm$ 0.06 & 99.12 $\pm$ 0.04 \\
		PFC \cite{pan2025overcoming} & \checkmark & 51.89 $\pm$ 0.04 & \underline{85.85 $\pm$ 0.05} & 32.42 $\pm$ 0.16 & \underline{99.17 $\pm$ 0.02} & \underline{90.77 $\pm$ 0.02} & 99.15 $\pm$ 0.03 \\
		\midrule
		MS-SHOT (ours) & \checkmark & \underline{79.21 $\pm$ 0.88} & \textbf{86.82 $\pm$ 1.43} & \textbf{64.19 $\pm$ 0.86} & \textbf{99.49 $\pm$ 0.28} & \textbf{91.07 $\pm$ 0.02} & \textbf{99.69 $\pm$ 0.02} \\
		
		\bottomrule
		\multicolumn{8}{l}{\footnotesize $\bullet$ The best outcome among all competitive methods is denoted in \textbf{bold}, while the second-best performance is indicated with \underline{underline}.}
	\end{tabular}
\end{table*}

\begin{table*}[htbp]
	\renewcommand\arraystretch{1}
	\centering
	\caption{\label{results_other_methods_hackrf}Classification accuracies (\%) comparison with existing methods on the HackRF dataset}
	\begin{tabular}{l c c c c c c c}
		\toprule
		Method & Source-data-free & h1 $\rightarrow$ h2 & h1 $\rightarrow$ h3 & h2 $\rightarrow$ h1 & h2 $\rightarrow$ h3 & h3 $\rightarrow$ h1 & h3 $\rightarrow$ h2\\
		\midrule
		Source only & & 47.48 $\pm$ 1.72 & 46.62 $\pm$ 1.62 & 51.95 $\pm$ 3.86 & 59.48 $\pm$ 9.74 & 52.79 $\pm$ 9.63 & 62.87 $\pm$ 9.97\\
		DANN \cite{ganin2016domain} & & 75.70 $\pm$ 2.02 & \textbf{73.99 $\pm$ 2.05} & 54.97 $\pm$ 9.27 & 70.77 $\pm$ 9.47 & 55.35 $\pm$ 1.91 & 76.47 $\pm$ 5.28\\
		MCD \cite{saito2018maximum} & & 73.59 $\pm$ 5.66 & 60.00 $\pm$ 5.07 & \underline{63.75 $\pm$ 9.38} & 73.70 $\pm$ 7.06  & 57.66 $\pm$ 5.20 & 65.57 $\pm$ 8.77\\
		SHOT \cite{liang2020we} & \checkmark & \underline{75.92 $\pm$ 2.98} & 52.47 $\pm$ 0.86 & 51.69 $\pm$ 1.88 & \underline{86.54 $\pm$ 1.72} & 54.64 $\pm$ 6.96 & \underline{78.23 $\pm$ 1.86}\\
		CoNMix \cite{kumar2023conmix} & \checkmark & 73.72 $\pm$ 1.32 & 55.35 $\pm$ 6.71 & 52.45 $\pm$ 0.93 & 78.30 $\pm$ 6.95 & 54.67 $\pm$ 1.63 & 52.03 $\pm$ 0.94\\
		GPUE \cite{litrico2023guiding} & \checkmark & 66.75 $\pm$ 4.17 & 55.28 $\pm$ 2.73 & 61.12 $\pm$ 1.52 & 78.18 $\pm$ 5.16 & \underline{65.70 $\pm$ 0.49} & 68.65 $\pm$ 2.51 \\
		UPA \cite{chen2024uncertainty} & \checkmark	& 69.33 $\pm$ 3.34 & 49.95 $\pm$ 0.06 & 55.32 $\pm$ 0.41 & 72.70 $\pm$ 2.48 & 57.70 $\pm$ 2.27 & 71.15 $\pm$ 0.04 \\
		SFADA \cite{HE2024110246} & \checkmark & 65.32 $\pm$ 1.06 & 53.38 $\pm$ 0.92 & 50.70 $\pm$ 3.19 & 76.12 $\pm$ 0.99 & 66.60 $\pm$ 0.69 & 72.25 $\pm$ 0.86 \\
		PFC \cite{pan2025overcoming} & \checkmark & 71.62 $\pm$ 3.64 & 65.62 $\pm$ 4.60 & 52.88 $\pm$ 3.25 & 66.32 $\pm$ 4.50 & 52.38 $\pm$ 2.62 & 69.49 $\pm$ 2.24 \\
		\midrule
		MS-SHOT (ours) & \checkmark & \textbf{77.34 $\pm$ 0.83} & \underline{72.29 $\pm$ 3.63} & \textbf{68.98 $\pm$ 2.01} & \textbf{92.99 $\pm$ 3.17} & \textbf{67.42 $\pm$ 0.95} & \textbf{78.49 $\pm$ 0.53} \\
		
		\bottomrule
	\end{tabular}
\end{table*}

\subsubsection{Implementation details}
The overall experimental setup is shown in Table~\ref{setup}.

The feature extractor for the Wisig dataset has been developed using a modified ResNet18 architecture~\cite{he2016deep}. It incorporates 1D convolution layers in place of the standard 2D convolution to better accommodate 1D time-series signals. The classifier consists of a fully connected layer.
The total computational complexity of our MS-SHOT model is approximately 130.7 million floating point operations (FLOPs), making it well-suited for lightweight deployment scenarios such as edge devices or real-time inference.

Following 7 epochs of source-domain training with a low learning rate, the model is saved to facilitate target domain adaptation. A relatively short source training phase is used to prevent overfitting and retain sufficient model flexibility for cross-domain adaptation.
The adaptation stage spans 20 epochs. The momentum parameter $\beta$ is set to 0.995. This choice is motivated by the observation that the class centers estimated from clustering all target-domain samples are more reliable than those computed from each individual mini-batch. To incorporate this prior knowledge, we adopt a momentum update strategy that assigns greater weight to the accumulated global statistics rather than the potentially noisy batch-wise estimates. Following the practice in prior works that utilize momentum-based center updates (e.g., in contrastive learning~\cite{he2020momentum}), a high momentum value close to one is commonly used to ensure stability and robustness. Therefore, we set $\beta = 0.995$ to strike a balance between responsiveness and consistency in updating the pseudo-label centers.

The temperature $\tau$ is set to 0.1 to sharpen the soft pseudo-label distribution and enhance discriminative learning. The weights $\lambda_1 = 0.3$, $\lambda_2 = 1.0$, and $\lambda_3 = 0.5$ in \eqref{eq:total} are chosen to balance the contributions of the entropy minimization, soft pseudo-labeling, and consistency regularization terms, respectively. The learning rate $\eta$ is set to 0.0006 to ensure stable convergence during adaptation.

To assess the model performance, we report the mean and standard deviation of the classification accuracy over the last 5 epochs to account for training stability and performance consistency.

\subsection{Experimental Results and Analysis}

This section presents a comprehensive evaluation of the proposed method under the source-free domain adaptation setting. We conduct a series of experiments to assess its performance, interpretability, and robustness across various conditions. First, we compare our method with several state-of-the-art approaches to validate its effectiveness. Then, we perform ablation studies to analyze the contributions of individual components. To further understand the behavior of the model, we evaluate the sensitivity of performance to key hyper-parameters, including the weights of loss terms and the momentum coefficient. We also visualize the learned feature representations using t-SNE to provide insights into class separability. Lastly, we examine the robustness of our method under varying signal-to-noise ratio (SNR) conditions and its scalability as the number of emitter classes increases.

Throughout all experiments in this section, we consider that the class distribution in the target domain is uniform and known, i.e., $\bm q = [N^t/K, \dots, N^t/K]$. The more challenging non-uniform and unknown $\bm q$ will be discussed in Section~\ref{sec:imbalance}.

\subsubsection{Comparison with Existing Methods} \label{sec:comparison}

We compare the MS-SHOT approach to existing methods, including DANN \cite{ganin2016domain}, SHOT \cite{liang2020we}, CoNMix \cite{kumar2023conmix},  MCD~\cite{saito2018maximum}, GPUE~\cite{litrico2023guiding}, UPA~\cite{chen2024uncertainty}, SFADA~\cite{HE2024110246}, and PFC~\cite{pan2025overcoming}. As a baseline, we also train a classification model only on source signals and test on target signals to benchmark performance. Our results, shown in Table \ref{results_other_methods}, reveal that all other methods outperform the source only model in SCRFFI. The last six columns of Table \ref{results_other_methods} provide the results of the SCRFFI experiment, which show that MS-SHOT outperforms alternative methods in most experiments. It is noteworthy that our method achieves a classification accuracy of 99.49\% for the ``$7-7\rightarrow8-8$'' task and 99.69\% for the ``$20-1\rightarrow7-14$'' task, comparable to the results of supervised training on the target domain. This may be attributed to the fingerprints of the two receivers being very similar, allowing a model trained in the source domain to accurately cluster target domain data based on emitter characteristics. The experimental results on the HackRF dataset are presented in the Table~\ref{results_other_methods_hackrf}. Our proposed method, MS-SHOT, outperforms existing methods, achieving improvements of 1.4\%, 5.2\%, 6.4\%, 1.7\%, and 0.2\% over the second-best method across various tasks. This demonstrates the effectiveness and superiority of MS-SHOT in the context of SCRFFI.
It is also worth commenting on the task 2-1 $\rightarrow$ 18-2, which shows both the lowest source-only accuracy (21.78\%) and the strongest performance gain after adaptation. This domain pair corresponds to the largest receiver-induced shift in our setting, so the target features are heavily distorted and strongly entangled, making it difficult for most baselines to avoid collapsing to a few dominant classes. By contrast, MS-SHOT leverages the nuclear-norm and class-prior regularization to discourage such class collapse and to maintain globally diverse predictions, while the momentum-based soft pseudo-labels provide a more stable refinement of target clusters. As a result, the relative improvement on 2-1 $\rightarrow$ 18-2 is particularly pronounced, even though its absolute accuracy remains lower than that of easier domain pairs.

\begin{table*}[htbp]
	\renewcommand\arraystretch{1}
	\centering
	\caption{\label{results_ablation}Classification accuracies (\%) of experiments for ablation studies}
	\begin{tabular}{c c c c c c c c c}
		\toprule
		\multicolumn{3}{c}{Components} & \multirow{2}{*}{14-7 $\rightarrow$ 3-19} & \multirow{2}{*}{1-1 $\rightarrow$ 1-19} & \multirow{2}{*}{2-1 $\rightarrow$ 18-2} & \multirow{2}{*}{7-7 $\rightarrow$ 8-8} & \multirow{2}{*}{2-19 $\rightarrow$ 19-2} &  \multirow{2}{*}{20-1 $\rightarrow$ 7-14}\\
		\cline{1-3}
		$\mathcal{L}_{nn}$ \& $\mathcal{L}_{\ell_1}$ & Soft Label & Momentum Center & ~ & ~ & ~ & ~ & ~ & ~ \\
		\midrule
		$\times$ & $\times$ & $\times$ & 33.26 $\pm$ 7.12 & 59.86 $\pm$ 5.03 & 21.78 $\pm$ 2.64 & 53.53 $\pm$ 1.70 & 62.91 $\pm$ 4.98 & 72.47 $\pm$ 6.13 \\
		\checkmark & $\times$ & $\times$ & 76.94 $\pm$ 0.81 & 84.36 $\pm$ 1.92 & \textbf{67.13 $\pm$ 0.94} & 98.52 $\pm$ 0.79 & 84.29 $\pm$ 0.03 & 87.83 $\pm$ 0.05 \\
		$\times$ & \checkmark & $\times$ & 41.94 $\pm$ 1.13 & 67.61 $\pm$ 0.25 & 16.67 $\pm$ 3.24 & 28.90 $\pm$ 0.66 & 70.02 $\pm$ 0.13 & 70.52 $\pm$ 1.52 \\
		\checkmark & \checkmark & $\times$ & 78.73 $\pm$ 0.33 & 85.44 $\pm$ 1.09 & 57.29 $\pm$ 3.02 & 99.25 $\pm$ 0.34 & 89.80 $\pm$ 0.07 & 99.29 $\pm$ 0.03 \\
		\midrule
		\checkmark & \checkmark & \checkmark & \textbf{79.21 $\pm$ 0.88} & \textbf{86.82 $\pm$ 1.43} & 64.19 $\pm$ 0.86 & \textbf{99.49 $\pm$ 0.28} & \textbf{91.07 $\pm$ 0.02} & \textbf{99.69 $\pm$ 0.02}\\
		
		\bottomrule
	\end{tabular}
\end{table*}

\begin{figure*} \centering
	\subfloat[]{\includegraphics[width=0.47\columnwidth]{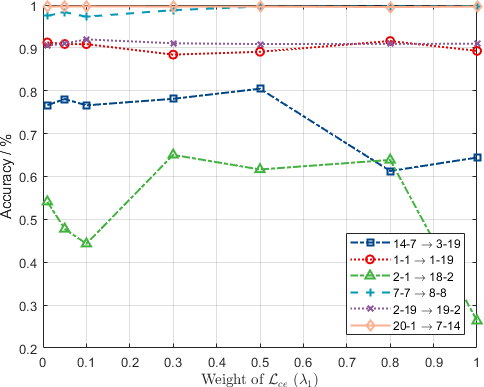}  
		\label{fig:weight_pseudo}  } \quad
	\subfloat[]{\includegraphics[width=0.47\columnwidth]{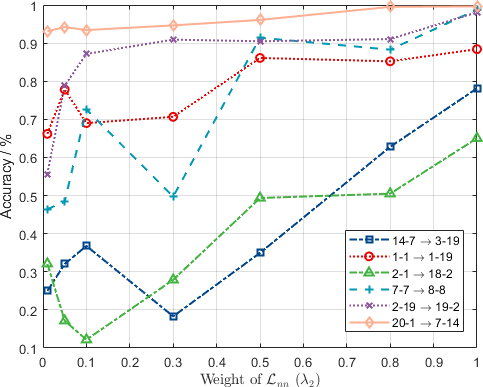}
		\label{fig:weight_nn}  } \quad
	\subfloat[]{\includegraphics[width=0.47\columnwidth]{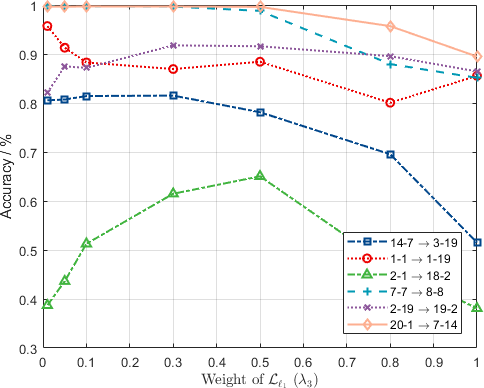}
		\label{fig:weight_1n} }
	\subfloat[]{\includegraphics[width=0.47\columnwidth]{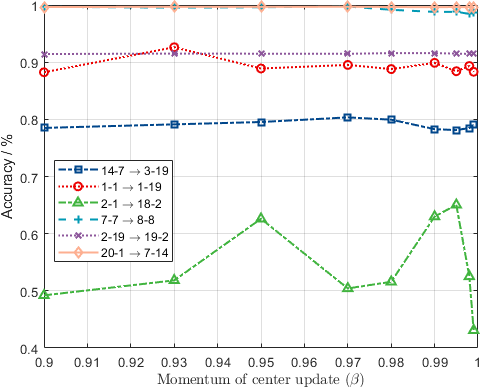}
		\label{fig:momentum} }
	\caption{{Experimental results of different hyper-parameters on model performance. (a) Weight of $\mathcal{L}_{ce}$ ($\lambda_1$). (b) Weight of $\mathcal{L}_{nn}$ ($\lambda_2$). (c) Weight of $\mathcal{L}_{\ell_1}$ ($\lambda_3$). (d) Momentum of the feature center update ($\beta$)}}
	\label{fig:weight}  
\end{figure*}

\subsubsection{Ablation Studies (Effect of $\mathcal{L}_{nn}$ and $\mathcal{L}_{\ell_1}$)}

We conduct an ablation study on each component of the MS-SHOT approach, demonstrating the effectiveness of each component. As depicted in Table \ref{results_ablation}, using both the two constraints $\mathcal{L}_{nn}$ and $\mathcal{L}_{\ell_1}$ significantly improves classification accuracy in the target domain. Specifically, in the SCRFFI experiments (the last 4 columns), only training with $\mathcal{L}_{nn}$ and $\mathcal{L}_{\ell_1}$ (second row) improves classification accuracy by at least 25\% compared to the source only approach (first row). In some experiments, its classification accuracy even approaches that of MS-SHOT (last row). In summary, leveraging $\mathcal{L}_{nn}$ and $\mathcal{L}_{\ell_1}$ can mitigate the cross-receiver effect without requiring source domain data. 

\subsubsection{Ablation Studies (Momentum Center-Guided Soft Pseudo-labeling)} 
In the ablation experiment, we divide MCSP into two parts, soft label, and momentum center. As depicted in Table \ref{results_ablation}, employing only soft labels (third row) can result in even worse results than the source only approach (first row). This could be due to relying heavily on the initial feature distribution and classification accuracy in the target domain for models trained on the source domain when using only soft labels. Without the component that aligns the feature distribution in both domains, relying purely on pseudo-labels (especially only soft labels) gets poor performance. On the other hand, soft labels can work well with other components to achieve better results. More specifically, in most experiments, using alignment and soft labels together (fourth row) performed better than only using $\mathcal{L}_{nn}$ and $\mathcal{L}_{\ell_1}$ (second row). 

%

\subsubsection{Sensitivity of hyper-parameters (Weight of losses)}
In our comprehensive SCRFFI experiments, we evaluate the impact of various hyper-parameters. Specifically, we assessed the effects of weights $\lambda_1, \lambda_2, \lambda_3$ in \eqref{eq:def_loss} on the experimental outcomes. Each hyperparameter was altered individually within the range of 0.01 to 1, while others remained constant, as the experimental setup. The results are shown in Fig.~\ref{fig:weight_pseudo}, \ref{fig:weight_nn}, \ref{fig:weight_1n}. For the weight $\lambda_1$ of pseudo-label supervision $\mathcal{L}_{ce}$, if $\lambda_1$ is too large, the model may over-fit the error information in the pseudo-labels. When $\lambda_1$ is too small, the model will ignore labeling information. An optimal $\lambda_1$ is critical to learning from label information while mitigating the noise impact due to the inherent error rate in pseudo-labels. As for the weight $\lambda_2$ for $\mathcal{L}_{nn}$, our experiments indicate the model performs best when it is set to 1 across all tasks. Regarding the weight $\lambda_3$ for $\mathcal{L}_{\ell_1}$, model performance remains stable within a range of 0.01 to 0.5; however, it deteriorates sharply when the value exceeds 0.5.

\subsubsection{Sensitivity of hyper-parameters (Momentum coefficient)}
We also investigate the impact of the Momentum coefficient in \eqref{eq:update_ck}, varying its value between 0.9 and 0.999, with results illustrated in Fig.~\ref{fig:momentum}. The model's performance is largely insensitive to changes in momentum, with most experiments showing stable results across the tested range. In specific cases, such as the ``$7-7 \rightarrow 8-8$'' experiment, an increased $\beta$ value (between 0.99 and 0.995) corresponds to improved performance. To balance the timeliness and stability of pseudo-labels, selecting a slightly higher $\beta$ value appears advantageous.

\begin{figure*} \centering 
	\subfloat[]{\includegraphics[width=0.45\columnwidth]{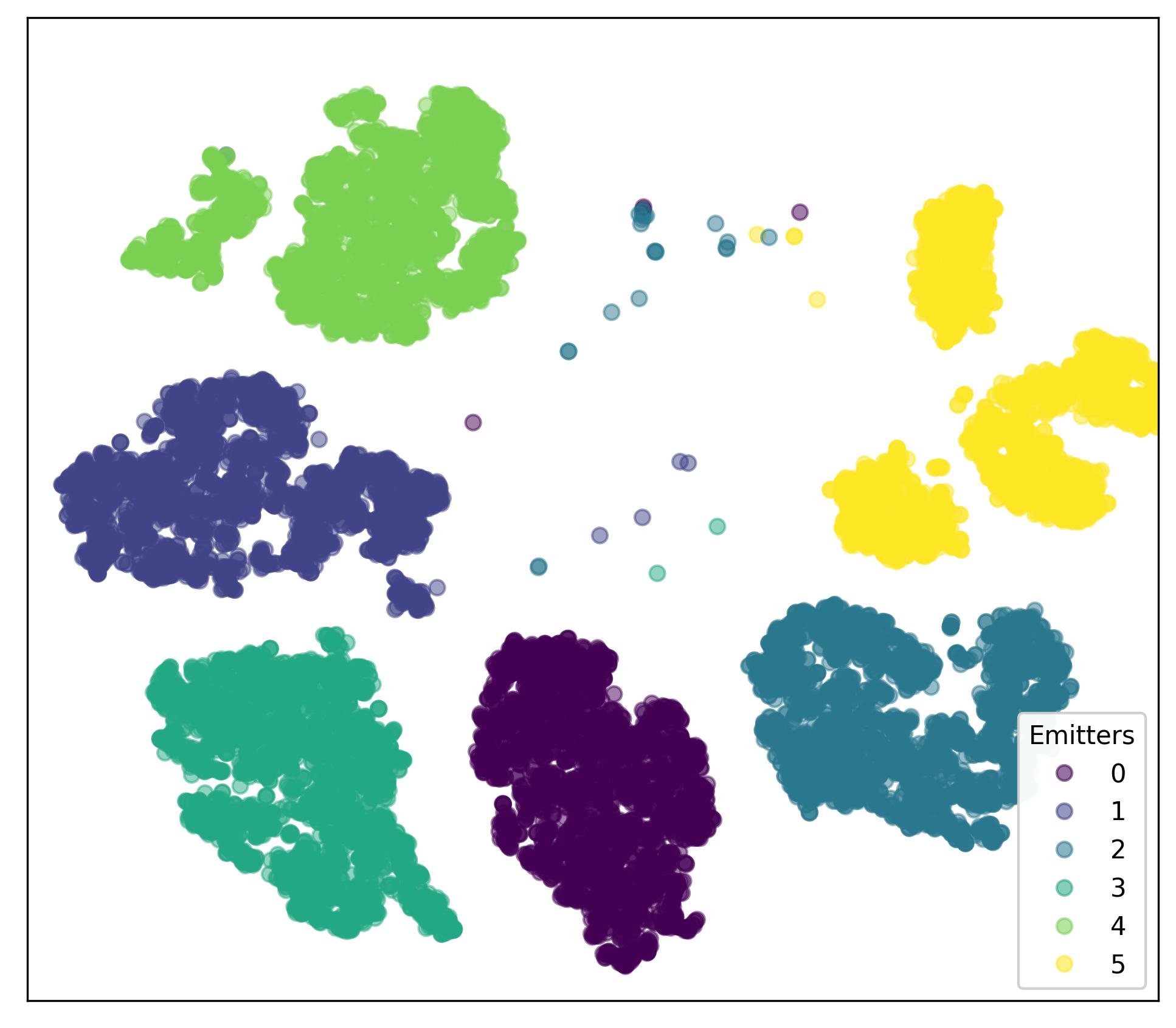}
		\label{fig:a}      } \quad
	\subfloat[]{\includegraphics[width=0.45\columnwidth]{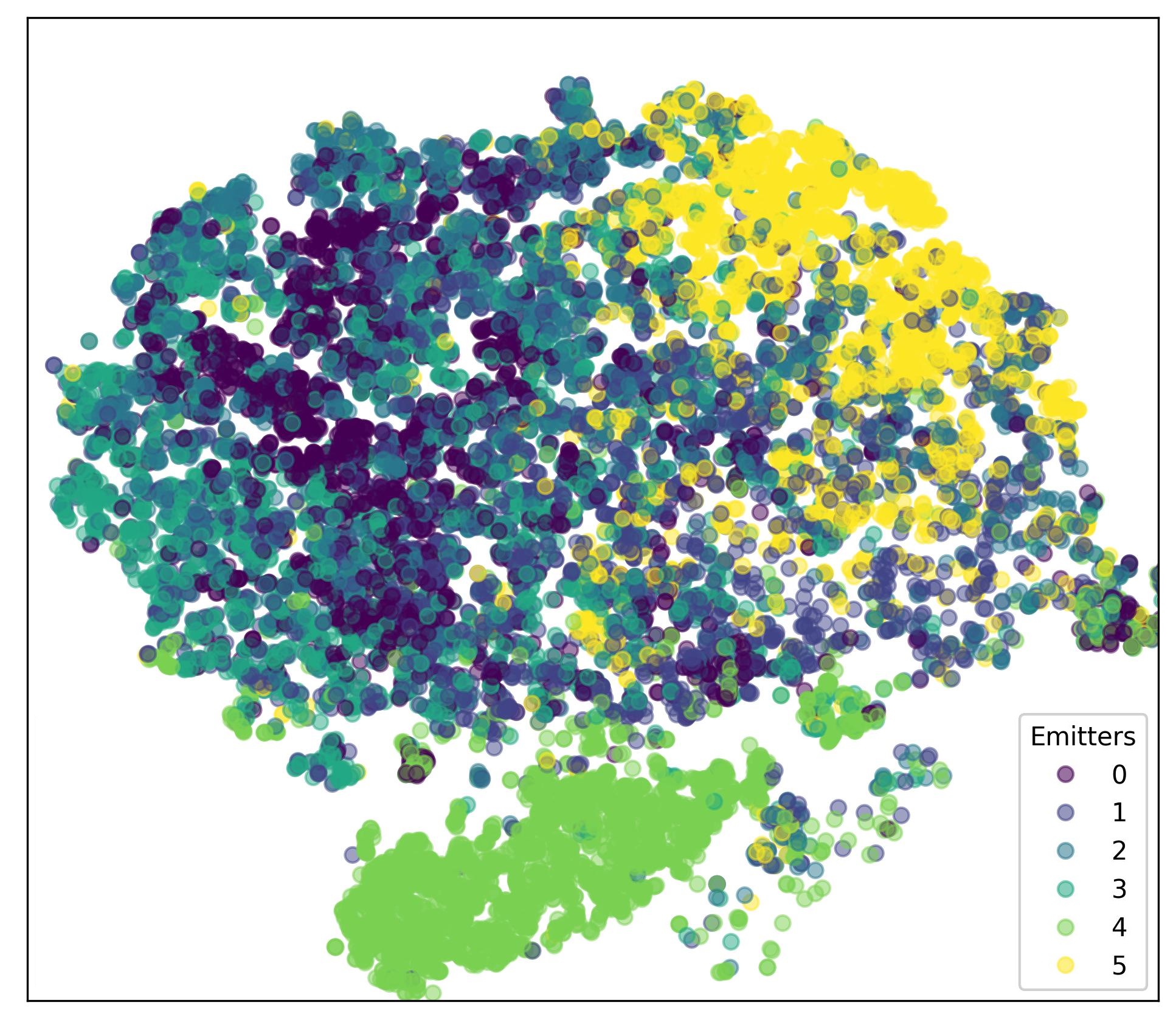}
		\label{fig:b}     } \quad
	\subfloat[]{\includegraphics[width=0.45\columnwidth]{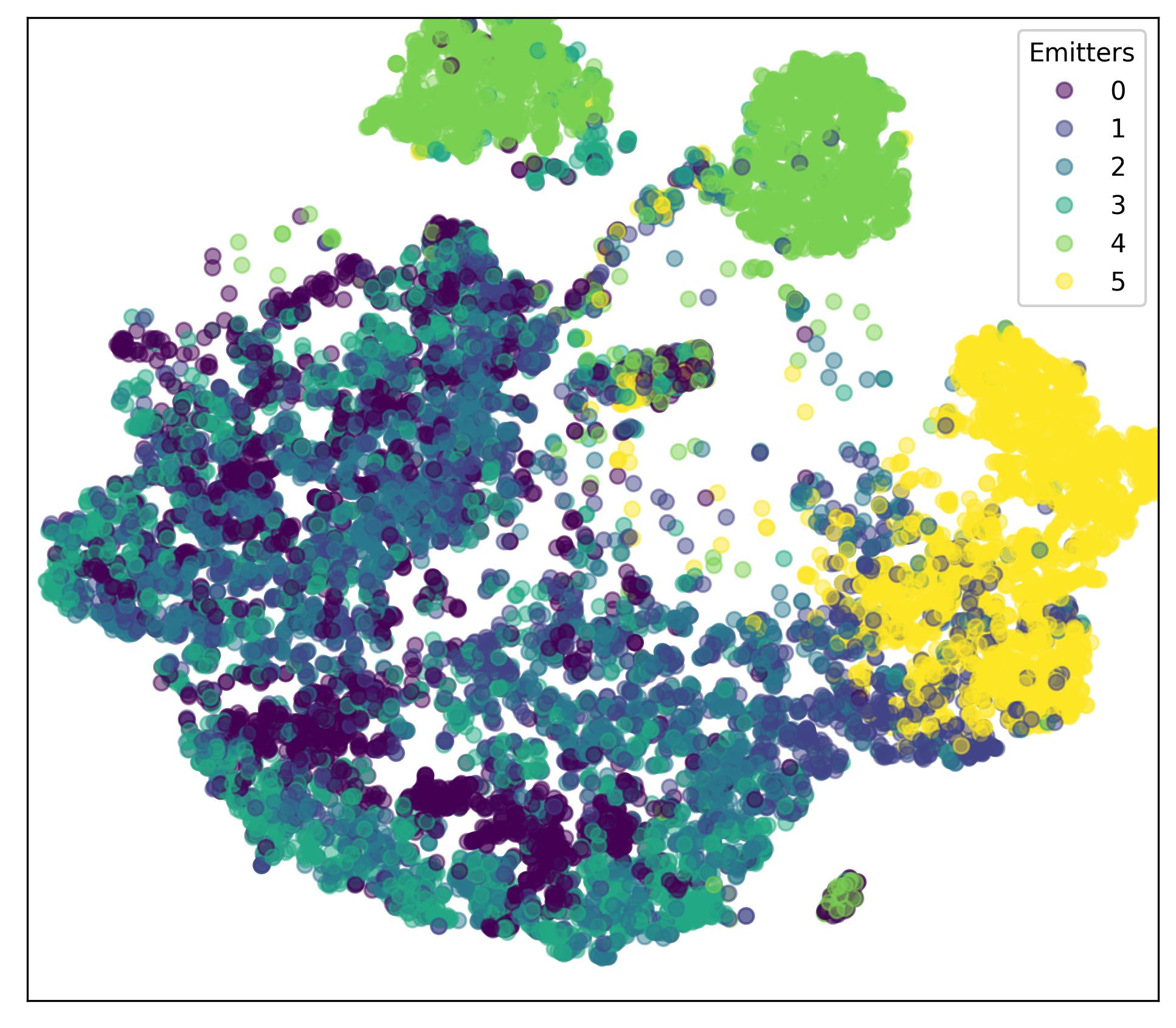}
			\label{fig:c}  } \quad
	\subfloat[]{\includegraphics[width=0.45\columnwidth]{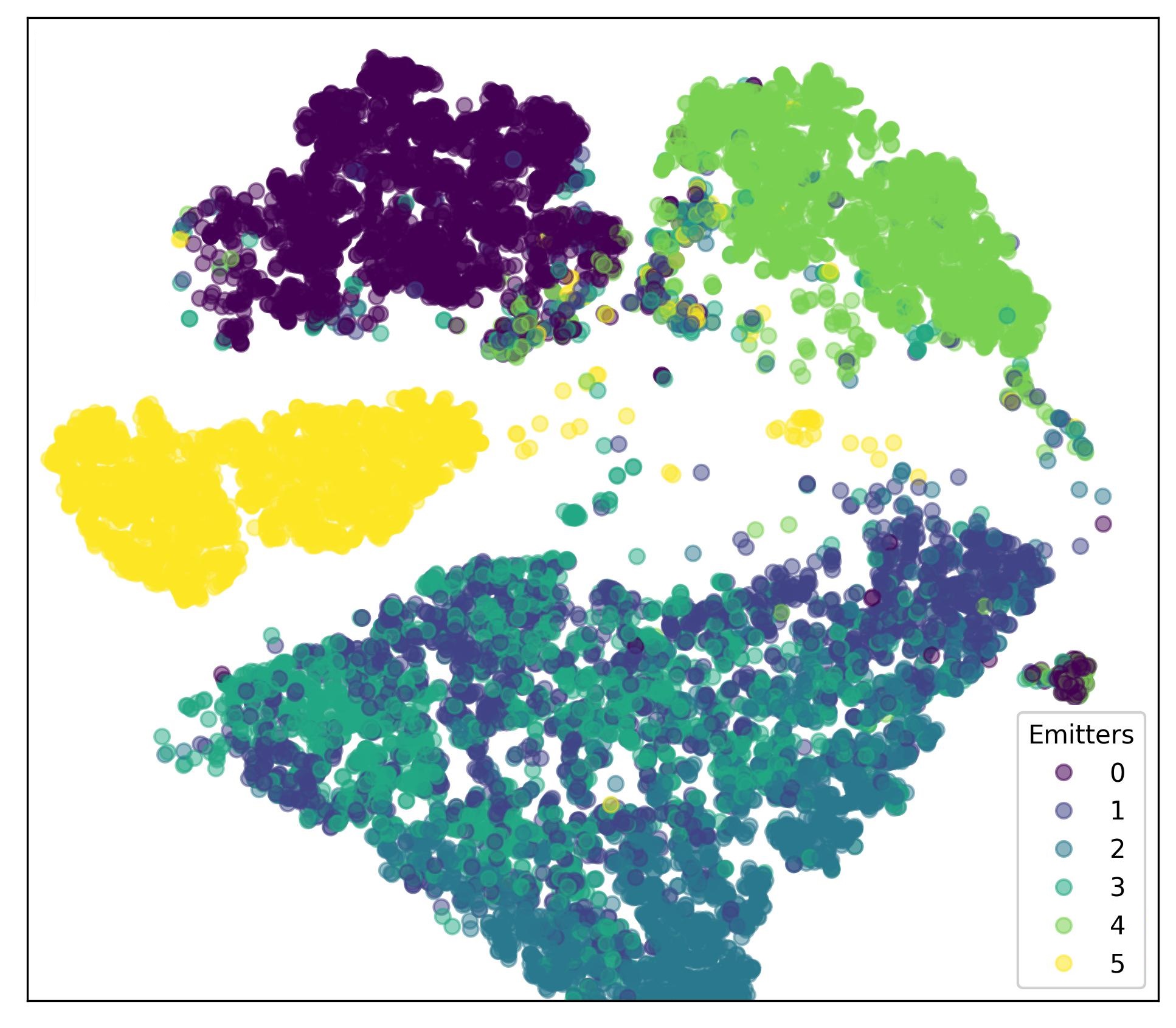}
	\label{fig:d} }
	\caption{The t-SNE visualization of (a) Source only learned representation in the source domain $\mathcal{S}$. (b) Source only learned representation in the target domain $\mathcal{T}$. (c) SHOT learned representation in the target domain $\mathcal{T}$. (d) MS-SHOT learned representation in the target domain $\mathcal{T}$.}
	\label{fig:tsne} 
\end{figure*}

\subsubsection{Feature Embedding Visualization with t-SNE}

In our study, we employed the t-distributed Stochastic Neighbour Embedding (t-SNE)~\cite{van2008visualizing} to visualize the two-dimensional projections of signal features extracted by the feature encoder $\phi$ across all emitters. This visualization aims to qualitatively assess the clustering behavior of features in the target domain and evaluate the effectiveness of different adaptation strategies. We conduct the analysis on the representative task ``$2-1 \rightarrow 18-2$'', comparing three models: source only, SHOT, and the proposed MS-SHOT.
While the source-only model performs well on the source domain—forming clear clusters for each emitter, it fails to generalize to the target domain. As observed in the projection of the source-only features on the target receiver, only a subset of emitters (e.g., classes 0, 4, and 5) retain well-separated clusters, whereas classes 1, 2, and 3 show significant overlap, indicating limited transferability of the learned representations. These classes likely share subtle and less discriminative signal characteristics, and their inter-class differences are further compressed by the nonlinear distortions introduced by the receiver hardware, which exacerbate the domain shift.
MS-SHOT introduces soft pseudo-label refinement and momentum-guided feature alignment to mitigate this shift. As visualized in the t-SNE plot, MS-SHOT significantly improves the feature separability over SHOT and the source-only model, especially for those emitter classes that retain some degree of latent separability. However, confusion remains among classes 1, 2, and 3. A closer analysis suggests that these classes were already entangled in the target domain even before adaptation, indicating that the issue lies not in the adaptation procedure per se, but in the insufficient discriminative structure of the target-domain features inherited from the source model. Since pseudo-label refinement relies on the quality of initial clustering, severely overlapping features in early stages may lead to limited adaptation gains.
This analysis highlights a key insight: while MS-SHOT is capable of adapting robust class boundaries in many cases, its performance is constrained when source representations lack sufficient separability in the target domain due to compounded distortions.
This limitation reflects the challenges of pseudo-label-driven methods under extreme domain shifts, and underscores the importance of future work in improving initial feature robustness or incorporating auxiliary guidance for ambiguous class structures.

\subsubsection{Robustness Evaluation under Varying SNR Conditions}
\begin{figure}
	\centerline{\includegraphics[scale=0.6]{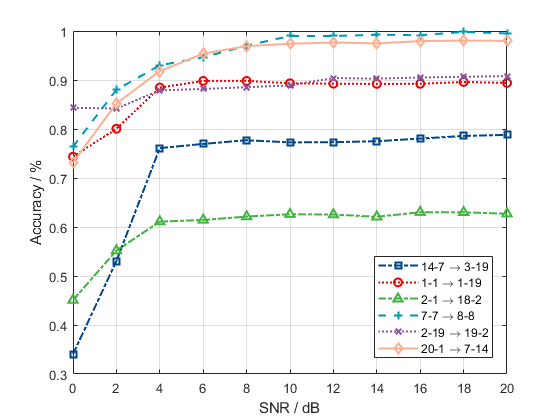}}
	\caption{The results of robustness evaluation under varying SNR conditions.}
	\label{fig:snr}
\end{figure}

To investigate the robustness of MS-SHOT under realistic noisy environments, we simulate signal degradation during both training and testing phases. In the source domain, Gaussian white noise is added to each training sample, with its SNR uniformly sampled from 0 dB to 20 dB. During inference, we further evaluate the adapted model under controlled target-domain SNR levels ranging from 0 dB to 20 dB in 2 dB increments.
The classification results under different SNR levels across six transfer tasks are summarized in Fig.~\ref{fig:snr}. Overall, MS-SHOT demonstrates strong robustness and graceful degradation in low-SNR conditions. For example, under extreme noise (SNR = 0 dB), the method still achieves 73.2\% and 76.4\% accuracy on the challenging tasks ``$20-1 \rightarrow 7-14$'' and ``$7-7 \rightarrow 8-8$'', respectively, and up to 84.3\% in ``$2-19 \rightarrow 19-2$''. As SNR increases, the performance steadily improves and saturates above 95\% for several tasks (e.g., ``$7-7 \rightarrow 8-8$'' and ``$20-1 \rightarrow 7-14$''), demonstrating the model's ability to exploit cleaner signal structure when available.
Interestingly, different transfer tasks exhibit varying degrees of SNR sensitivity. Tasks such as ``$1-1 \rightarrow 1-19$'' and ``$2-19 \rightarrow 19-2$'' show relatively minor fluctuations across SNRs, suggesting lower domain discrepancy and better initial pseudo-label quality. In contrast, ``$2-1 \rightarrow 18-2$'' and ``$14-7 \rightarrow 3-19$'' exhibit significant degradation at low SNRs (e.g., 45.1\% and 34.0\% at 0 dB, respectively), reflecting either greater cross-receiver shift or intrinsically more confusable emitter classes. These results highlight that while MS-SHOT is robust overall, the effectiveness of adaptation remains influenced by both channel quality and underlying domain alignment.
This experiment validates the practical feasibility of MS-SHOT in dynamic wireless environments with fluctuating SNRs.

\subsubsection{Scalability Evaluation with Varying Transmitter Classes}
\begin{figure}
	\centerline{\includegraphics[scale=0.6]{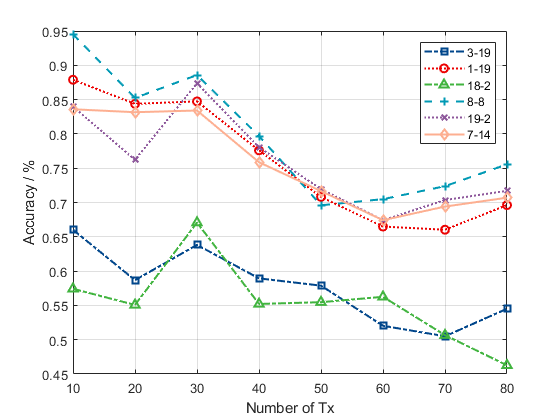}}
	\caption{The results of scalability evaluation with varying emitter classes.}
	\label{fig:Tx}
\end{figure}

To further evaluate the scalability and practicality of MS-SHOT under large-scale deployment settings, we construct an expanded dataset from the original Wisig collection. Specifically, we extract signals from 80 Tx and 26 receivers over four consecutive days of collection. For each Tx–Rx pair, 150 signal samples are used. We designate 80\% of the data for training and 20\% for testing. Among the 26 receivers, we fix the 6 target receivers used in Section~\ref{sec:comparison}, and treat the remaining 20 receivers as source receivers. For each adaptation task, the model is trained using all four days of source-domain data from the 20 receivers and adapted on the corresponding target receiver using target domain data. The number of Tx classes (i.e., the number of distinct Tx) is then varied from 10 to 80 in steps of 10 to assess the model’s robustness against increasing classification complexity.
As shown in Fig.~\ref{fig:Tx}, MS-SHOT exhibits favorable scalability behavior. In the low-to-medium scale range (10 to 30 Tx), classification accuracy remains high and stable across most transfer pairs. For example, the ``$8-8$'' and ``$1-19$'' adaptation tasks consistently maintain over 85\% accuracy, indicating that MS-SHOT can effectively adapt even when a moderate number of new device identities are introduced.
As the number of Tx classes increases beyond 40, a gradual performance decline is observed. This degradation is expected, as more classes introduce higher inter-class confusion and greater demands on the feature discriminability of the model. Nonetheless, the performance drop is not abrupt. For instance, in the ``$7-14$'' adaptation task, accuracy decreases smoothly from 83.59\% (10 classes) to 70.70\% (80 classes), showing that MS-SHOT still retains effective adaptation capacity under increased complexity. Moreover, even in the most challenging scenarios, such as ``$18-2$'' adaptation, the accuracy remains above random guessing and shows resilience to scale increases, despite the weaker domain alignment in this pair.
Another notable observation is that certain adaptation pairs (e.g., $3-19$ and $18-2$) show more fluctuation across different class scales. This could be attributed to variations in domain shift severity and class separability, especially in receiver combinations with higher hardware asymmetry. These results suggest that while scale poses challenges to adaptation, MS-SHOT remains robust and can generalize across varying data sizes and Tx population scales.
In summary, this set of experiments demonstrates that MS-SHOT is capable of adapting under both moderate and large-scale Tx scenarios, maintaining competitive accuracy and stable behavior. Its ability to operate across datasets of increasing complexity makes it well-suited for real-world RFFI deployments where the number of devices may be large and continuously growing.

\subsection{Performance under Non-uniform and unknown class distribution} \label{sec:imbalance}

To further evaluate the effectiveness of our method in more realistic scenarios, we consider the case where the target domain exhibits an non-uniform label distribution. 
Specifically, we use the same Wisig dataset as in previous experiments, but artificially control the proportion of samples in each class of the target domain to simulate imbalance. In this experiment, the number of samples in each target-domain class is set to 30\%, 45\%, 60\%, 75\%, 90\%, and 100\% of the corresponding class sample counts used in the previous uniform setting, resulting in a target-domain class ratio of approximately 0.3:0.45:0.6:0.75:0.9:1.

We compare our method with several baseline algorithms. In addition, we consider three variants of our proposed MS-SHOT method:
\begin{itemize}
	\item MS-SHOT (w/ known): Class priors in the target domain are known.
	\item MS-SHOT (w/ uniform): Class priors are unknown but assumed to be uniform.
	\item MS-SHOT (w/ estimate): Class priors are unknown and estimated from the pseudo-labeling function $\tilde{f}^t$.
\end{itemize}

\begin{table*}[htbp]
	\renewcommand\arraystretch{1}
	\centering
	\caption{\label{tab:imbalance_results}Classification accuracies (\%) under non-uniform and unknown target domain setting on the Wisig dataset}
	\begin{tabular}{l c c c c c c c}
		\toprule
		Method & Source-data-free & 14-7 $\rightarrow$ 3-19 & 1-1 $\rightarrow$ 1-19 & 2-1 $\rightarrow$ 18-2 & 7-7 $\rightarrow$ 8-8 & 2-19 $\rightarrow$ 19-2 & 20-1 $\rightarrow$ 7-14\\
		\midrule
		DANN \cite{ganin2016domain} & & 48.77 & 74.71 & 20.10 & 69.67 & 78.52 & 96.44 \\
		MCD \cite{saito2018maximum} & & 49.92 & \underline{81.38} & 33.02 & 70.92 & 75.81 & 82.98 \\
		SHOT \cite{liang2020we} & \checkmark & 70.52 & 55.44 & 25.12 & 75.17 & 65.92 & 99.03 \\
		CoNMix \cite{kumar2023conmix} & \checkmark & 35.17 & 50.21 & 24.62 & 74.21 & 61.15 & 75.31 \\
		GPUE \cite{litrico2023guiding} & \checkmark & 43.16 & 64.19 & 39.26 & 67.22 & 69.82 & 96.65 \\
		UPA \cite{chen2024uncertainty} & \checkmark & 53.12 & 81.21 & 38.60 & 73.54 & 54.58 & 94.31 \\
		SFADA \cite{HE2024110246} & \checkmark & 57.10 & \textbf{82.23} & 20.33 & 65.69 & 65.92 & 99.12 \\
		PFC \cite{pan2025overcoming} & \checkmark & 52.31 & 64.96 & 35.81 & 69.71 & 55.75 & 72.96 \\
		\midrule
		MS-SHOT (w/ known) & \checkmark & \underline{73.96} & 71.71 & \textbf{46.08} & \textbf{99.79} & \underline{80.58} & \textbf{99.73} \\
		MS-SHOT (w/ uniform) & \checkmark & 65.69 & 61.94 & 37.75 & 80.90 & 74.40 & 88.48 \\
		MS-SHOT (w/ estimate) & \checkmark & \textbf{76.17} & 78.06 & \underline{41.62} & \underline{88.79} & \textbf{88.94} & \underline{99.65} \\
		\bottomrule
	\end{tabular}
\end{table*}

As shown in Table~\ref{tab:imbalance_results}, our proposed MS-SHOT (w/ estimate) consistently outperforms its uniform counterpart MS-SHOT (w/ uniform) across all tasks. This highlights the importance of adapting to the actual class distribution rather than assuming uniformity, especially in realistic, non-uniform settings.

More importantly, despite not having access to the true target class prior, MS-SHOT (w/ estimate) achieves performance comparable to the idealized setting MS-SHOT (w/ known) where the class distribution is fully known. In some tasks such as $14-7 \rightarrow 3-19$ and $2-19 \rightarrow 19-2$, the estimated variant even surpasses the known-prior variant. These results clearly demonstrate the effectiveness of our pseudo-label-based estimation strategy in recovering meaningful class-level statistics, enabling the model to approach the optimal performance achievable under full prior knowledge.

We further note that, on some domain pairs, MS-SHOT (w/ estimate) slightly outperforms MS-SHOT (w/ known) despite not using the true class prior. This is because the class-proportion constraint is enforced on predictions and pseudo-labels, which can be noisy under strong domain shift. Strictly matching the true prior may then over-emphasize confusing classes and reinforce label noise, mildly hurting top-1 accuracy. In contrast, MS-SHOT (w/ estimate) derives its prior $\hat{\bm q}$ from the current pseudo-label histogram, so that the global constraint is better aligned with the model's confidence pattern; since we evaluate overall accuracy rather than class-balanced accuracy, this self-consistent adjustment can occasionally yield slightly better performance than the known-prior variant.

Overall, this experiment confirms that MS-SHOT (w/ estimate) is not only robust to class imbalance but also capable of self-correcting for unknown label shifts, making it a strong and practical solution for source-free adaptation in real-world scenarios.

\section{Conclusion and discussion}\label{sec:conclusion}

In this paper, we addressed the critical challenge of ensuring robust RF fingerprint identification (RFFI) in edge intelligence environments, where receiver variability and resource constraints pose significant obstacles. We identified the limitations of existing deep learning-based RFFI methods, which often fail to consider the impact of receiver hardware differences, leading to degraded performance when models are deployed across heterogeneous receivers. To address this issue, we introduced the source-data-free cross-receiver RFFI (SCRFFI) problem and proposed the Momentum Soft pseudo-label Source Hypothesis Transfer (MS-SHOT) method. MS-SHOT enables effective model adaptation to new receivers using only unlabeled signals from the target receiver, without requiring access to source data.

Although MS-SHOT demonstrates robust performance across diverse cross-receiver settings, its effectiveness depends on the target-domain features being at least partially separable. When the domain discrepancy is very large, for example due to strong receiver-induced distortions, the extracted target features may become highly entangled and lose class discriminability, limiting the adaptation effect. This reflects a common limitation shared by most source-free domain adaptation methods. For security-critical applications, we recommend complementing MS-SHOT with auxiliary safeguards such as runtime confidence monitoring or human-in-the-loop validation to ensure reliable deployment in challenging environments.

Moreover, although our experiments focus on a single-source receiver, the proposed two-stage pipeline (supervised pre-training on labeled source data followed by source-data-free adaptation on the target receiver) is naturally extensible to multi-source scenarios. In practice, labeled data from several source receivers can be aggregated, or combined via federated training, to obtain a stronger source hypothesis that is then adapted to new deployment receivers using the same MS-SHOT mechanism.

In future work, it is worthwhile to explore hybrid adaptation strategies that integrate a small amount of labeled target data or domain-specific priors to improve robustness under extreme domain shift, as well as to systematically study such multi-source/multi-receiver extensions in practical SCRFFI deployments.

\bibliographystyle{IEEEtran}
\bibliography{reference}

\vfill

\section{Supplementary Material}
\setcounter{equation}{0}

\renewcommand\theequation{A\arabic{equation}} 
\subsection{Proof of Theorem~\ref{thm:1}}\label{sec:appendix-thm1}
The proof consists of two parts. 
We first prove that the (unknown) ground-truth labeling function $f^t$ is feasible for problem \eqref{eq:main_known}; we then exploit the triangle inequality and VC dimension theory to derive the desired result.

Let the ground-truth label function $f^t \colon \mathcal{X} \to \Delta_K$ assign each target sample $x_i \in \mathcal{T}$ its ground-truth label $y_i = f^t(x_i)$. Construct the matrix $\bm Q_{f^t} \in \{0,1\}^{N^t \times K}$ such that each row is a one-hot vector:
\[
[\bm Q_{f^t}]_{i,k} =
\begin{cases}
	1, & \text{if } f^t(x_i) = k, \\
	0, & \text{otherwise}.
\end{cases}
\]

Let $n_k = N^t \alpha_k$ denote the number of samples with label $k$ under the known class prior $\bm \alpha$. Then we have
\[
\bm 1_{N^t}^{\!\top} \bm Q_{f^t} = [n_1, n_2, \dots, n_K] = N^t \bm \alpha,
\]
which implies that the constraint~\eqref{eq:main_c_case1} is satisfied with
\[
\left\| \bm 1_{N^t}^{\!\top} \bm Q_{f^t} - N^t \bm \alpha \right\|_1 = 0 \leq \gamma.
\]

Next, note that the matrix $\bm Q_{f^t}$ has $n_k$ rows equal to the $k$-th standard basis vector. Thus, the nuclear norm of $\bm Q_{f^t}$ satisfies
\[
\|\bm Q_{f^t}\|_* = \sum_{k=1}^K \sqrt{n_k},
\]
which exactly equals $\zeta(\bm \alpha)$ defined in \eqref{eq:main_b_case1}, and the constraint~\eqref{eq:main_b_case1} is also satisfied.

Therefore, $f^t$ satisfies both constraints, implying that $f^t$ is a feasible solution to the problem~\eqref{eq:main_known}. It guarantees that the optimization problem is well-posed: it admits at least one solution, namely the (unknown) ground-truth $f^{t}$.

Next, we introduce the    triangle inequality lemma for classification error:
For any $h_\ell \in {\cal H},~ \ell=1,2,3$, the following inequality holds true.
\begin{equation}\label{eq:triangle_ineq}
	\epsilon(h_1,h_2) \leq \epsilon(h_1,h_3)  + \epsilon(h_3,h_2) 
\end{equation}
With the triangle inequality for classification error, we are ready to complete the whole proof.
\begin{align*}
	&\epsilon^t(\hat{h}, f^t) \\
	\leq & \hat{\epsilon}^t(\hat{h}, f^t) + {c_1}/{2}  \\
	\leq & \hat{\epsilon}^t(\hat{h}, \tilde{f}^t) + \hat{\epsilon}^t(\tilde{f}^t, f^t) + {c_1}/{2}  \\
	\leq & \hat{\epsilon}^t(f^t, \tilde{f}^t) + \hat{\epsilon}^t(\tilde{f}^t, f^t) +  {c_1}/{2} \\
	= & 2\hat{\epsilon}^t(f^t, \tilde{f}^t)  +  {c_1}/{2}  \\
	\leq &  2 {\epsilon}^t(f^t, \tilde{f}^t)  + c_1 
\end{align*}
where  the first inequality follows from a standard application of the VC theory~\cite{vapnik1998statistical} to bound the expected risk $\epsilon^t(\hat{h}, f^t)$ by its empirical estimate $\hat{\epsilon}^t(\hat{h}, f^t)$. Namely, if ${\cal T}$ is an $N^t$-size  i.i.d. sample, then with probability exceeding $1-\rho$, 
\[\epsilon^t(\hat{h}, f^t) \leq  \hat{\epsilon}^t(\hat{h}, f^t) + \sqrt{\frac{d(\log(2N^t/d)+1) + \log(4/\rho)}{N^t}}.\]	The second inequality is due to the triangle inequality for classification error~\cite{ben2010theory}, the third inequality is because $\hat{h}$ is optimal for the problem in \eqref{eq:main} and $f^t$ is a feasible solution, and the last inequality is again due to the VC theory. This completes the proof. 

\subsection{Proof of Theorem~\ref{thm:2}}\label{sec:appendix-tm2}
Define $h^\star$ as the ideal joint hypothesis on both domains, i.e. 
\[ h^\star = \arg\min_{h\in \cal H} \epsilon^t(h, f^t) +  \epsilon^t(h, f^s) \] and let $\eta^\star$ be the corresponding minimum value.  By setting  $\tilde{f}^t = h^s$ in~\eqref{eq:Thm1}, we have
\begin{equation}\label{eq:Thm2}
	\begin{aligned}
		& \epsilon^t(\hat{h}, f^t)\\
		\leq &  2\epsilon^t(h^s, f^t) + c_1   \\
		\leq & 2(\epsilon^t(h^s, h^\star) + \epsilon^t(h^\star, f^t) ) + c_1 \\
		\leq & 2 (\epsilon^t(h^\star, f^t) +  \epsilon^s(h^s, h^\star) +   |\epsilon^t(h^s, h^\star) -  \epsilon^s(h^s, h^\star) |) + c_1 \\
		\leq & 2 (\epsilon^t(h^\star, f^t) +  \epsilon^s(h^s, h^\star)) +   d_{\cal H}({\cal D}^s, {\cal D}^t) + c_1 \\
		\leq & 2 (\epsilon^t(h^\star, f^t) + \epsilon^s(h^\star, f^s) + \epsilon^s(h^s, f^s) ) +  d_{\cal H}({\cal D}^s, {\cal D}^t) + c_1 \\
		= & 2 \epsilon^s(h^s, f^s) + 2\eta^\star + d_{\cal H}({\cal D}^s, {\cal D}^t) + c_1
	\end{aligned}
\end{equation}
where we have used the triangle inequality in Section\ref{sec:appendix-thm1}. This completes the proof.
\subsection{Proof of Corollary~\ref{corrollay}}\label{sec:corrollay}
From~\eqref{eq:Thm1}, we have $\epsilon^t(\hat{h}, f^t) \leq 2\epsilon^t(\tilde{f}^t, f^t) + c_1$, which together with  $\epsilon^t(\tilde{f}^t, f^t)< \epsilon^t(h^s, f^t)$ implies $\epsilon^t(\hat{h}, f^t) < 2\epsilon^t(h^s, f^t) + c_1$. Notice that the upper bound can be further bounded by $2 \epsilon^s({h}^s, f^s) + c_2$ (cf. the right-hand side of the first inequality in~\eqref{eq:Thm2}), and thus the  inequality in~\eqref{eq:corrollay} holds.

\subsection{Proof of Theorem~\ref{thm:unknown}}\label{sec:appendix-thm3}
We aim to show that the unknown label function $f^t$ is a feasible solution to the optimization problem under the estimated target class prior $\hat{\bm{q}}$, i.e., $\bm Q_{f^t}$ satisfies both regularization constraints.

Let $\bm Q_{f^t}$ denote the one-hot label matrix corresponding to the true (but unknown) label function $f^t$. Since $f^t$ is one-hot, each row of $\bm Q_{f^t}$ contains a single 1, and the column sums represent the number of samples assigned to each class by $f^t$, denoted as $\bm q^t = \bm 1_{N^t}^{\top} \bm Q_{f^t}$.

\textbf{(1) Verification of the $\ell_1$ constraint:}

By Assumption~\ref{assum:distribution}, we are given that the pseudo-label class prior $\hat{\bm q}$ is $\gamma$-close to the true class prior:
\[
\bigl\|\bm 1_{N^{t}}^{\!\top}\bm Q_{f^{t}} - \hat{\bm q}\bigr\|_{1} \le \gamma.
\]
This directly implies that $\bm Q_{f^t}$ satisfies the constraint.

\textbf{(2) Verification of the nuclear norm constraint:}

Since $\bm Q_{f^t}$ is a one-hot matrix, its nuclear norm equals the $\ell_2$-norm of its column counts:
\[
\|\bm Q_{f^t}\|_* = \sum_{k=1}^K \sqrt{n_k},
\]
where $n_k$ is the number of samples in class $k$ under $f^t$.

From Assumption~\ref{assum:distribution}, and the inequality $\| \bm q - \hat{\bm q} \|_1 \le \gamma$, it follows that for each class $k$,
\[
|n_k - \hat{n}_k| \le \gamma \quad \Rightarrow \quad n_k \ge \hat{n}_k - \gamma.
\]

Using the inequality $\sqrt{a} \ge \sqrt{b} - \dfrac{|a - b|}{2\sqrt{b}}$ for $a, b > 0$ and applying it to each $n_k$ and $\hat{n}_k$ yields:
\[
\sqrt{n_k} \ge \sqrt{\hat{n}_k} - \frac{\gamma}{2\sqrt{\hat{n}_k}}.
\]
Summing over $k$ gives:
\begin{align}
	\|\bm Q_{f^t}\|_* = \sum_{k=1}^K \sqrt{n_k} 
	&\ge \sum_{k=1}^K \left( \sqrt{\hat{n}_k} - \frac{\gamma}{2\sqrt{\hat{n}_k}} \right) \notag \\
	&\ge \sum_{k=1}^K \left(\sqrt{\hat{n}_k} - \frac{\gamma}{2\sqrt{\hat{n}_{\min}}}\right),
\end{align}
where $\hat{n}_{\min} = \min_k \hat{n}_k$. According to the definition of $\zeta(\hat{\bm q})$ in constraint~\eqref{eq:main_b_case2},  under the assumption $\gamma < 2\hat{n}_{\min}$, it follows that:
\[
\|\bm Q_{f^t}\|_* \ge \zeta(\hat{\bm q}) > 0.
\]
Therefore, $\bm Q_{f^t}$ satisfies both constraints, and the function $f^t$ is a feasible solution to the optimization problem under the estimated class prior in the target domain.

Moreover, by following the same argument as in Theorem~\ref{thm:1}, Theorem~\ref{thm:2}, and Corollary~\ref{corrollay}, we can show that the generalization bound for $\hat h$ remains valid in this setting since the structure of the optimization problem and the application of VC-dimension-based results do not depend on whether the class prior is known or estimated.

\end{document}